\documentclass[12pt]{article}
\usepackage{amsmath}
\usepackage{graphicx,psfrag,epsf}
\usepackage{enumerate}
\usepackage{natbib}
\usepackage{url} 

\newcommand{\blind}{0}

\usepackage{graphics}
\usepackage{mathbbol}
\usepackage{amssymb}
\usepackage{hyperref}
\usepackage{bm}
\usepackage{color}
\usepackage{amsthm}
\usepackage{amsfonts}
\usepackage{subfig}
\usepackage{float}
\usepackage{comment}
\usepackage{srctex}

\newtheorem{theorem}{Theorem}

\definecolor{JP}{RGB}{255,0,0}
\newcommand{\JP}[1]{\textcolor{JP}{[J: #1]}}

\definecolor{JY}{RGB}{255,0,0}

\graphicspath{{images/}}

\begin{document}

\def\spacingset#1{\renewcommand{\baselinestretch}%
{#1}\small\normalsize} \spacingset{1}


\if0\blind
{
  \title{\bf Estimating Time-Varying Graphical Models}
  \author{Jilei Yang \thanks{The authors gratefully acknowledge the following support: 
  		UCD Dissertation Year Fellowship (JLY), NIH 1R01EB021707 (JLY and JP) and NSF-DMS-1148643 (JP).
  	}\\
    Department of Statistics, University of California,  Davis\\
    and \\
    Jie Peng\footnote{Correspondence author: jiepeng@ucdavis.edu}\\
    Department of Statistics, University of California,  Davis}
  \maketitle
} \fi

\if1\blind
{
  \bigskip
  \bigskip
  \bigskip
  \begin{center}
    {\LARGE\bf Estimating Time-Varying Graphical Models}
\end{center}
  \medskip
} \fi

\bigskip
\begin{abstract}

In this paper, we study time-varying graphical models  based on data measured over a temporal grid.  
Such models are motivated by the needs to describe and understand evolving interacting relationships among a set of  random variables in many real applications, for instance  the study of how stocks interact with each other and how such interactions change over time. 


We propose  a new model,  \textit{LOcal Group Graphical Lasso Estimation} (\texttt{loggle}), under the assumption that  the graph topology changes gradually over time.  Specifically, \texttt{loggle} uses a novel local group-lasso type penalty to efficiently incorporate information from neighboring time points and to impose structural smoothness of the graphs. We implement  an ADMM based   algorithm  to fit the \texttt{loggle} model. This algorithm utilizes blockwise fast computation and  pseudo-likelihood approximation to improve computational efficiency. An R package \texttt{loggle} has also been developed.   

We evaluate the performance of \texttt{loggle} by simulation experiments. We also apply  \texttt{loggle} to S\&P 500 stock price data and demonstrate that \texttt{loggle} is able to reveal the interacting  relationships among stocks and  among industrial sectors in a time period that covers the  recent global financial crisis. 

\end{abstract}

\noindent%
{\it Keywords:} ADMM algorithm,  Gaussian graphical model, group-lasso,   pseudo-likelihood approximation, S\&P 500
\vfill

\newpage
\spacingset{1.45} 

\section{Introduction}

In recent years, there are many problems where the study of the interacting relationships among a large number of variables is of interest. One  popular approach is to characterize interactions as conditional dependencies: Two variables are interacting with each other if they are conditionally dependent given the rest of the variables. An advantage of using conditional dependency instead of marginal dependency (e.g., through correlation) is that we are aiming  for more direct interactions after taking out the effects of  the  rest of the variables.   Moreover, if the random variables follow a multivariate normal distribution, then the elements of the inverse covariance matrix $\bm{\Sigma}^{-1}$ (a.k.a. precision matrix) would indicate  the presence/absence of such interactions. This is because under normality, two variables are conditionally dependent given the rest of the variables if and only if the corresponding element of the precision matrix is nonzero. Furthermore, we can  represent such interactions  by a graph $G=(V,E)$, where the node set $V$ represents the random variables of interest and the edge set $E$ consists of pairs $\{i,j\}$ where the $(i,j)$th element of $\bm{\Sigma}^{-1}$ is nonzero. Such models are referred to as \textit{Gaussian graphical models (GGM)}.



Many methods have been proposed to learn GGMs   when the number of variables is large (relative to the sample size), including  \cite{meinshausen2006high}, \cite{yuan2007model}, \cite{friedman2008sparse},   \cite{banerjee2008model}, \cite{rothman2008sparse}, \cite{peng2009partial}, \cite{lam2009sparsistency}, \cite{ravikumar2011high}, \cite{cai2011constrained}.  These methods rely on the sparsity assumption, i.e., only a small subset of  elements in the precision matrix is nonzero, to deal with challenges posed by  high-dimension-low-sample-size.

The aforementioned methods learn a single  graph based on the observed data.  However, when data are observed over a temporal or spatial grid,  the underlying graphs might change over time/space. For example, the  relationships among stocks  could  evolve over time as illustrated by Figure \ref{fig:graph_loggle}. If we had described them by a single graph, the results would be misleading. This necessitates the
study of time-varying graphical models.  

When the graphs/covariance matrices change over time, the observations would not be identically distributed anymore. To deal with this challenge,  one approach  is to assume  the  covariance matrices change smoothly over time. For example, in \cite{zhou2010time}, \cite{song2009keller}, \cite{kolar2010estimating}, \cite{kolar2012estimating}, \cite{wang2014inference}, \cite{monti2014estimating}, \cite{gibberd2014high},  \cite{gibberd2017regularized}, kernel estimates of the  covariance matrices are used  in the objective functions. However, smoothness of the covariance matrix alone does  not tell us how the  graph topology would evolve over time, despite in 
practice this is often of more interests than estimating the covariance matrices. Moreover, imposing certain assumption on how the graph topology changes over time could greatly facilitate interpretation  and consequently provide insights about the interacting relationships.


One type of time-varying graphical models utilize fused-lasso type penalties \citep{ahmed2009recovering, kolar2010estimating, kolar2012estimating, monti2014estimating, gibberd2014high, wit2015inferring, gibberd2017regularized, hallac2017network}, such that the estimated  graph topology would be  piecewise constant.  This is particularly convenient when we are primarily interested in detecting  jump points and abrupt changes.


In this paper, we consider  the \textit{structural smoothness} assumption, which assumes  that the graph topology is gradually changing over time.  For this purpose, we propose  \textit{LOcal Group Graphical Lasso Estimation} (\texttt{loggle}), a novel  time-varying graphical model that imposes structural smoothness through a local group-lasso type penalty. 
The \texttt{loggle} method is able to efficiently utilize neighborhood information and  is also able to adapt to the local degree of smoothness  in a data driven fashion. Consequently, the \texttt{loggle} method is flexible and effective  for a wide range of scenarios including models with both time-varying and time-invariant graphs. Moreover, we implement an ADMM based  algorithm that utilizes blockwise fast computation and pseudo-likelihood approximation to achieve computational efficiency and we use cross-validation to select tuning parameters.  We demonstrate the  performance of \texttt{loggle} through  a simulation study. Finally, we  apply \texttt{loggle} to the S\&P 500 stock price data to reveal how interactions among  stocks and  among industrial  sectors evolve during the recent global financial crisis. An R package \texttt{loggle} has also been developed. 

The rest of the paper is organized as follows. In  Section \ref{sec:method}, we introduce the \texttt{loggle} model, model fitting algorithms and strategies for model tuning. In Section \ref{sec:simulation}, we present simulation results to demonstrate the performance of \texttt{loggle} and compare it with two existing methods. We report the application on S\&P 500 stock price data in Section \ref{sec:real_data}, followed by conclusions  in Section \ref{sec:discussion}. Technical details are in an Appendix.  Additional details are deferred to a Supplementary Material.

\section{Methods}\label{sec:method}

\subsection{Local Group Graphical Lasso Estimation}\label{subsec:loggle}
In this section, we introduce \texttt{loggle} \textit{(LOcal Group Graphical Lasso Estimation)} for  time-varying graphical models. 


Let  $\bm{X}(t)=(X^1(t),\ldots,X^p(t))^T\sim \mathcal{N}_p(\bm{\mu}(t),\bm{\Sigma}(t))$ be a $p$-dimensional Gaussian random vector indexed by $t \in [0,1]$.
We assume $\bm{X}(t)$'s are independent across $t$. We also assume the mean function $\bm{\mu}(t)$ and the covariance function $\bm{\Sigma}(t)$ are smooth in $t$. We denote the observations by  $\{\bm{x}_k\}_{k \in \mathcal{I}}$ where $\mathcal{I}=\{1,\cdots, N\}$, $\bm{x}_k$ is a realization of   $\bm{X}({t_k})$ ($k \in \mathcal{I}$) and $0\leq t_1\leq\ldots\leq t_N\leq1$.  
For simplicity, we assume that the observations are centered so that  $\bm{x}_k$ is drawn  from $\mathcal{N}_p(\bm{0},\bm{\Sigma}(t_k))$. In practice, we can achieve this by  subtracting the estimated mean $\hat{\bm{\mu}}(t_k)$  from  $\bm{x}_k$. See Section \ref{sec:detrend} of the Supplementary Material for details.

Our goal is to estimate  the precision matrix $\bm{\Omega}(t):=\bm{\Sigma}^{-1}(t)$ based on the observed data  $\{\bm{x}_k\}_{k \in \mathcal{I}}$ and then construct the edge set (equiv. the graph topology) $E(t)$ based on the sparsity pattern of the estimated precision matrix $\hat{\bm{\Omega}}(t)$.
We further assume that the edge set (equiv. the graph topology) changes gradually over time.


To estimate the precision matrix $\bm{\Omega}(t_k)$ at the $k$th observed time point, we propose to minimize a locally weighted  negative log-likelihood function with a local group-lasso penalty: 
\begin{equation}
\label{eq:loggle}
L(\mathbb{\Omega}_k):=\frac{1}{\sqrt{|\mathcal{N}_{k,d}|}}\sum_{i\in \mathcal{N}_{k,d}}\left[\text{tr}\left(\bm{\Omega}(t_i)\hat{\bm{\Sigma}}(t_i)\right)-\log|\bm{\Omega}(t_i)|\right]
+\lambda\sum_{u\neq v}\sqrt{\sum_{i\in \mathcal{N}_{k,d}}\Omega_{uv}(t_i)^2},
\end{equation}
where $\mathcal{N}_{k,d}=\{i \in \mathcal{I}:|t_i-t_k|\leq d\}$ denotes the indices of the time points centered around $t_k$ with neighborhood width $d$ and $|\mathcal{N}_{k,d}|$ denotes the number of elements in $\mathcal{N}_{k,d}$;  $\mathbb{\Omega}_k=\{\bm{\Omega}(t_i)\}_{i\in \mathcal{N}_{k,d}}$ denotes the set of precision matrices  within this neighborhood  and $\Omega_{uv}(t_i)$ denotes the $(u,v)$-th element of $\bm{\Omega}(t_i)$;  $\hat{\bm{\Sigma}}(t)=\sum_{j=1}^N\omega_h^{t_j}(t)\bm{x}_j\bm{x}_j^T$ is the kernel estimate of the covariance matrix at time $t$, where the weights $\omega_h^{t_j}(t)=\frac{K_h(t_j-t)}{\sum_{j=1}^NK_h(t_j-t)}$, $K_h(\cdot)=K(\cdot/h)$ is a symmetric nonnegative kernel function and $h(>0)$ is the  bandwidth. 

We obtain: 
$$\hat{\mathbb{\Omega}}_k=\{\hat{\bm{\Omega}}(t_i)\}_{i\in \mathcal{N}_{k,d}}=\arg\min_{\bm{\Omega}(t_i)\succ\bm{0},\;i\in \mathcal{N}_{k,d}} L(\mathbb{\Omega}_k),$$
and set $\hat{\bm{\Omega}}(t_k)$ as the estimated precision matrix at time $t_k$. 
Since the group-lasso type penalty is likely to over-shrink the elements in the precision matrix, we further perform model refitting by maximizing the weighted log-likelihood function under the constraint of  the estimated edge set (equiv. sparsity pattern). We denote the refitted estimate by 
$\hat{\bm{\Omega}}^\text{rf}(t_k)$.  
Note that, the precision matrix may be estimated at any time point $t\in[0,1]$: If $t \notin \{t_k: k \in \mathcal{I}\}$, then choose an integer $\tilde{k} \notin \mathcal{I}$ and define $t_{\tilde{k}}=t$, $\mathcal{\tilde{I}}=\mathcal{I} \cup \{\tilde{k}\}$ and $\mathcal{N}_{\tilde{k},d}=\{i \in \mathcal{\tilde{I}}:|t_i-t_{\tilde{k}}|\leq d\}$. For  simplicity of exposition,  throughout we  describe the \texttt{loggle}  fits at observed time points. 
 

The use of the kernel estimate $\hat{\bm{\Sigma}}(t)$  
is justified by the assumption that the covariance matrix  $\bm{\Sigma}(t)$  is smooth in $t$. This allows us to borrow information from neighboring  time points.   In practice, we often replace the kernel smoothed covariance matrices by kernel smoothed correlation matrices which amounts to data standardization.

The penalty term $\lambda\sum_{u\neq v}\sqrt{\sum_{i\in \mathcal{N}_{k,d}}\Omega_{uv}(t_i)^2}$ is a group-lasso type sparse regularizer \citep{yuan2006model, danaher2014joint} that  makes the graph topology change smoothly over time. The degree of such smoothness is controlled by the tuning parameter $d (>0)$, the larger the neighborhood width $d$, the more gradually the graph topology would change.  The overall sparsity of the graphs is controlled by the tuning parameter $\lambda (>0)$, the larger the $\lambda$,  the sparser the graphs tend to be. The factor $\frac{1}{\sqrt{|\mathcal{N}_{k,d}|}}$ in equation (\ref{eq:loggle}) is to make $\lambda$  comparable for different $d$.

The \texttt{loggle} model includes two existing time-varying graphical models as special cases. Specifically,  in \cite{zhou2010time},  $\bm{\Omega}(t_k)$ is estimated by minimizing a weighted negative log-likelihood function with a lasso penalty:
$$\min_{\bm{\Omega}(t_k)\succ\bm{0}}\text{tr}\left(\bm{\Omega}(t_k)\hat{\bm{\Sigma}}(t_k)\right)-\log|\bm{\Omega}(t_k)|+\lambda\sum_{u\neq v}|\Omega_{uv}(t_k)|,$$ which is a special case of \texttt{loggle} by setting $d=0$.  This method utilizes the smoothness of the  covariance matrix by introducing the kernel estimate $\hat{\bm{\Sigma}}(t)$ in the likelihood function. However, it ignores potential structural smoothness of the graphs and thus might not utilize the data most efficiently. Hereafter, we refer to this method as \texttt{kernel}. 

On the other hand, \cite{wang2014inference} propose  to use a (global) group-lasso penalty to estimate $\bm{\Omega}(t_k)$'s simultaneously:
$$\min_{\{\bm{\Omega}(t_k)\succ\bm{0}\}_{k=1,\ldots,N}}\sum_{k=1}^N\left[\text{tr}\left(\bm{\Omega}(t_k)\hat{\bm{\Sigma}}(t_k)\right)-\log|\bm{\Omega}(t_k)|\right]
+\lambda\sum_{u\neq v}\sqrt{\sum_{k=1}^N\Omega_{uv}(t_k)^2}.$$ This is  another special case of \texttt{loggle} by setting $d$ large enough to cover the entire time interval $[0, 1]$ (e.g., $d=1$). The (global) group-lasso penalty  makes the estimated precision matrices have the same sparsity pattern (equiv. same graph topology) across the entire time domain. This could be too  restrictive for many applications where the graph topology is expected to change over time. Hereafter, we refer to this method as \texttt{invar}.


\subsection{Model Fitting}\label{subsec:algorithm}

Minimizing the objective function in equation (\ref{eq:loggle}) with respect to $\mathbb{\Omega}_k$ is a convex optimization problem. This can be solved by an ADMM (\textit{alternating directions method of multipliers}) algorithm (See details in Section \ref{sec:ADMM} of the Supplementary Material). ADMM algorithms can converge to the global optimum for convex optimization problems under very mild conditions. A comprehensive introduction can be found in \cite{boyd2011distributed}. However, this ADMM algorithm  involves  $|\mathcal{N}_{k,d}|$ eigen-decompositions of $p\times p$ matrices (each corresponding to a time point in the neighborhood)  in every iteration,  which is computationally very expensive when $p$ is large. In the following, we  propose a fast blockwise algorithm as well as pseudo-likelihood type approximation of the objective function to speed up the computation.



\subsubsection*{Fast blockwise algorithm}
If the solution is block diagonal (after suitable permutation of the variables), then we can apply the ADMM algorithm to each block separately, and consequently reduce the computational complexity from $O(p^3)$ to $\sum_{l=1}^LO(p_l^3)$, where $p_l$'s are the block sizes and $\sum_{l=1}^Lp_l=p$. 

We establish the following theorems when there are two blocks. These results follow similar results in \cite{witten2011new} and \cite{danaher2014joint} and  can be easily extended to an arbitrary number of blocks. 
\begin{theorem}\label{thm:block_diag}
	Suppose the solution of minimization of  (\ref{eq:loggle}) with respect to $\mathbb{\Omega}_k$ has the following form (after appropriate variable permutation): 
	$$\hat{\bm{\Omega}}(t_i)=\left(\begin{array}{c c}\hat{\bm{\Omega}}_1(t_i) & \bm{0} \\ \bm{0} &\hat{\bm{\Omega}}_2(t_i)\end{array}\right), ~~~ i\in \mathcal{N}_{k,d}, $$
	where all $\hat{\bm{\Omega}}_1(t_i)$'s have the same dimension. Then $\{\hat{\bm{\Omega}}_1(t_i)\}_{i\in \mathcal{N}_{k,d}}$ and $\{\hat{\bm{\Omega}}_2(t_i)\}_{i\in \mathcal{N}_{k,d}}$ can be obtained by minimizing (\ref{eq:loggle})  on  the respective sets of variables separately.
\end{theorem}
\begin{theorem}\label{thm:disconnect}
	Let $\{G_1,G_2\}$ be a non-overlapping partition of the $p$ variables. A necessary and sufficient condition for the variables in $G_1$ to be completely disconnected from those in $G_2$ in all  estimated precision matrices $\{\hat{\bm{\Omega}}(t_i)\}_{i \in \mathcal{N}_{k,d}}$ through minimizing   (\ref{eq:loggle}) is:
	$$\frac{1}{|\mathcal{N}_{k,d}|}\sum_{i\in \mathcal{N}_{k,d}}\hat{\Sigma}_{uv}(t_i)^2\leq\lambda^2, ~~~~ \text{for all} ~~ u\in G_1,\;v\in G_2.$$
	
\end{theorem}
The proof of Theorem \ref{thm:block_diag} is straightforward through inspecting the Karush-Kuhn-Tucker (KKT) condition of the optimization problem of minimizing (\ref{eq:loggle}). The proof of Theorem \ref{thm:disconnect} is given in  Appendix  \ref{sec:disconnect}.


Based on Theorem \ref{thm:disconnect}, we propose the following \textit{fast blockwise ADMM algorithm}: 
\begin{enumerate}[(i)]
	\item
	Create a $p\times p$ adjacency matrix $\bm{A}$. For $1\leq u\neq v\leq p$, set the off-diagonal elements  $A_{uv}=0$  if $\frac{1}{|\mathcal{N}_{k,d}|}\sum_{i\in \mathcal{N}_{k,d}}\hat{\Sigma}_{uv}(t_i)^2\leq\lambda^2$; and $A_{uv}=1$, if otherwise.
	\item
	Identify the connected components, $G_1,\cdots,G_L$, given the adjancency matrix $\bm{A}$. Denote their  sizes by $p_1,\cdots,p_L$ ($\sum_{l=1}^Lp_l=p$).
	\item For $l=1,\cdots,L$,  if $p_l=1$, i.e., $G_l$ contains only one variable, say the $u$th variable, then set $\hat{\Omega}_{uu}(t_i)=(\hat{\Sigma}_{uu}(t_i))^{-1}$ for $i\in \mathcal{N}_{k,d}$; If $p_l>1$, then apply the ADMM algorithm to the $p_l$ variables in $G_l$ to obtain the corresponding $\{\hat{\bm{\Omega}}_l(t_i)\}_{i \in \mathcal{N}_{k,d}}$. 
\end{enumerate}

\subsubsection*{Pseudo-likelihood approximation}\label{sec:pseudo_likelihood}
Even with the  fast blockwise algorithm, the computational cost could  still be high due to the eigen-decompositions. In the following,  we propose a \textit{pseudo-likelihood approximation} to speed up step (iii) of this algorithm. In practice, this approximation has been able to reduce computational cost by as much as 90\%. For simplicity of exposition, the description is based on the entire set of the variables.

The proposed approximation is based on the following well known fact that relates the elements of the precision matrix to the coefficients of regressing one variable to the rest of the variables \citep{meinshausen2006high, peng2009partial}. Suppose a random vector $(X^1,\ldots,X^p)^T$ has mean zero and covariance matrix $\bm{\Sigma}$. Denote the precision matrix by  $\bm{\Omega}=((\Omega_{uv})):=\bm{\Sigma}^{-1}$. If we write $X^u=\sum_{v\neq u}\beta_{uv}X^v+\epsilon_u$, where the residual $\epsilon_u$ is uncorrelated with $\{X^v:v\neq u\}$, then  $\beta_{uv}=-\frac{\Omega_{uv}}{\Omega_{uu}}$. 
Note that, $\beta_{uv}=0$ if and only if $\Omega_{uv}=0$.  Therefore identifying the sparsity pattern of the precision matrix is equivalent to identifying sparsity pattern of the regression coefficients. 


We consider minimizing the following local group-lasso penalized weighted $L_2$ loss function for estimating $\bm{\beta}(t_k)=(\beta_{uv}(t_k))_{u\neq v}$:
\begin{equation}\label{eq:pseudo_naive}
L_{PL}(\mathbb{B}_k):=\frac{1}{\sqrt{|\mathcal{N}_{k,d}|}}\sum_{i\in \mathcal{N}_{k,d}}\left[\frac{1}{2}\sum_{u=1}^p||\bm{X}_u-\sum_{v\neq u}\beta_{uv}(t_i)\bm{X}_v||_{\bm{W}_h(t_i)}^2\right]
+\lambda\sum_{u\neq v}\sqrt{\sum_{i\in \mathcal{N}_{k,d}}\beta_{uv}(t_i)^2}, 
\end{equation}
where $\mathbb{B}_k=\{\bm{\beta}(t_i)\}_{i\in \mathcal{N}_{k,d}}$ is the set of $\bm{\beta}(t_i)$'s within the neighborhood centered around $t_k$ with neighborhood width $d$; $\bm{X}_u=(x^u_1,\ldots,x^u_N)^T$ is the sequence of the $u$th variable in observations $\{\bm{x}_j\}_{1\leq j\leq N}$ and $\bm{W}_h(t_i)={\rm diag}\{\omega_h^{t_i}(t_j)\}_{1\leq j\leq N}$ is a weight matrix. The $\bm{W}$-norm of a vector $\bm{z}$ is defined as $||\bm{z}||_{\bm{W}}=\sqrt{\bm{z}^T\bm{W}\bm{z}}$. Once $\hat{\bm{\beta}}(t_k)$ is obtained through minimizing (\ref{eq:pseudo_naive}) with respect to $\mathbb{B}_k$, we can derive the estimated edge set at $t_k$: $\widehat{E}(t_k)=\{\{u,v\}: \hat{\beta}_{uv}(t_k)\neq0,  ~~ u < v\}$.

The objective function (\ref{eq:pseudo_naive})  may be viewed as an approximation of the likelihood based objective function (\ref{eq:loggle}) through the aforementioned regression connection by ignoring the correlation among the residuals $\epsilon_u$'s.  We refer to this approximation as the \textit{pseudo-likelihood approximation}. 
However,  minimizing (\ref{eq:pseudo_naive}) cannot guarantee symmetry of edge selection, i.e., $\hat{\beta}_{uv}(t)$ and $\hat{\beta}_{vu}(t)$ being simultaneously  zero or nonzero. To achieve this, we modify (\ref{eq:pseudo_naive}) by using a \textit{paired group-lasso penalty}  \citep{friedman2010applications}:
\begin{equation}\label{eq:pseudo}
\begin{split}
\tilde{L}_{PL}(\mathbb{B}_k)=&\frac{1}{\sqrt{|\mathcal{N}_{k,d}|}}\sum_{i\in \mathcal{N}_{k,d}}\left[\frac{1}{2}\sum_{u=1}^p||\bm{X}_u-\sum_{v\neq u}\beta_{uv}(t_i)\bm{X}_v||_{\bm{W}_h(t_i)}^2\right]\\
&+\lambda\sum_{u<v}\sqrt{\sum_{i\in \mathcal{N}_{k,d}}\left[\beta_{uv}(t_i)^2+\beta_{vu}(t_i)^2\right]}.
\end{split}
\end{equation}
The paired group-lasso penalty guarantees simultaneous selection of   $\beta_{uv}(t)$ and $\beta_{vu}(t)$. 

The objective function (\ref{eq:pseudo}) can be rewritten as: 
\begin{equation}\label{eq:pseudo_concat}
\tilde{L}_{PL}(\mathbb{B}_k)=\frac{1}{\sqrt{|\mathcal{N}_{k,d}|}}\sum_{i\in \mathcal{N}_{k,d}}\frac{1}{2}||\mathbf{Y}(t_i)-\mathbf{X}(t_i)\bm{\beta}(t_i)||_2^2+\lambda\sum_{u<v}\sqrt{\sum_{i\in \mathcal{N}_{k,d}}\left[\beta_{uv}(t_i)^2+\beta_{vu}(t_i)^2\right]},
\end{equation}
where $\mathbf{Y}(t_i)=(\tilde{\bm{X}}_1(t_i)^T,\ldots,\tilde{\bm{X}}_p(t_i)^T)^T$ is  an $Np\times1$ vector with $\tilde{\bm{X}}_u(t_i)=\sqrt{\bm{W}_h(t_i)}\bm{X}_u$ being an $N\times1$ vector ($u=1,\cdots, p$);  $\mathbf{X}(t_i)=(\tilde{\bm{X}}_{(1,2)}(t_i),\ldots,\tilde{\bm{X}}_{(p,p-1)}(t_i))$ is an $Np\times p(p-1)$ matrix, with $\tilde{\bm{X}}_{(u,v)}(t_i)=(\bm{0}_N^T,\ldots,\bm{0}_N^T,\tilde{\bm{X}}_v(t_i)^T,\bm{0}_N^T,\ldots,\bm{0}_N^T)^T$ being  an $Np\times1$ vector, where $\tilde{\bm{X}}_v(t_i)$ is in the $u$th block ($1\leq u\neq v \leq p$); and $\bm{\beta}(t_i)=(\beta_{12}(t_i),\ldots,\beta_{p,p-1}(t_i))^T$ is a $p(p-1)\times1$ vector.

We implement an ADMM algorithm to minimize (\ref{eq:pseudo_concat}), which does not involve eigen-decomposition and thus is much faster than the ADMM algorithm for minimizing the original likelihood based  objective function (\ref{eq:loggle}). 	This is because, the $L_2$ loss used in (\ref{eq:pseudo})  and (\ref{eq:pseudo_concat}) is quadratic in the parameters $\mathbb{B}_k$  as opposed  to the negative log-likelihood loss used in (\ref{eq:loggle}) which has a log-determinant term.  Moreover,  $\mathbf{X}(t_i)$ is actually a block diagonal matrix: $\mathbf{X}(t_i)={\rm diag}\{\tilde{\mathbf{X}}_{(-u)}(t_i)\}_{1\leq u\leq p}$, where $\tilde{\mathbf{X}}_{(-u)}(t_i)=(\tilde{\bm{X}}_1(t_i),\ldots,\tilde{\bm{X}}_{u-1}(t_i), \tilde{\bm{X}}_{u+1}(t_i),\ldots,\tilde{\bm{X}}_p(t_i))$ is an $N\times(p-1)$ matrix.  Therefore, computations can be done in a blockwise fashion and potentially can be parallelized. The detailed algorithm is given in Appendix \ref{sec:ADMM_PL}.

\subsection{Model Tuning}\label{subsec:tuning_parameter}
In the \texttt{loggle} model, there are three  tuning parameters, namely, $h$ -- the kernel bandwidth (for $\hat{\bm \Sigma}(t)$'s), $d$ -- the neighborhood width (for $\mathcal{N}_{k,d}$'s) and $\lambda$ -- the sparsity parameter. In the following, we describe $V$-fold cross-validation (CV) to choose these parameters.

Recall that observations are made on a temporal grid. So we create the validation sets by including every $V$th data point and the corresponding training set would be the rest of the data points. E.g.,  for $V=5$, the 1st validation set would include observations at $t_1,t_6,t_{11}, \cdots$, the 2nd validation set would include those at $t_2, t_7, t_{12}, \cdots$, etc.  In the following, let $\mathcal{I}_{(v)}$ denote the indices of the time points in the $v$th validation set and $\mathcal{I}_{-(v)}$ denote those in the $v$th training set ($v=1,\cdots, V$). 

Let $h_{\rm grid}, d_{\rm grid}, \lambda_{\rm grid}$ denote the tuning grids from which $h$, $d$ and $\lambda$, respectively, are chosen.  See Section \ref{sec:simulation} for an example of the tuning grids. We recommend to choose $d$ and $\lambda$  separately for each $t_k$ as the degrees of sparsity and  smoothness of the graph topology may vary over time.  On the other hand, we recommend to choose a common $h$ for all time points.  


Given time $t_k$ and $h$, for $(d_k,\lambda_k)$, we obtain the refitted estimate $\hat{\bm{\Omega}}^\text{rf}_{-(v)}(t_k; d_k,\lambda_k,h)$ by applying \texttt{loggle} to the $v$th training set $\{\bm{x}_i\}_{i\in \mathcal{I}_{-(v)}}$ ($v=1,\cdots, V$). As mentioned in Section \ref{subsec:loggle},  this can be done even if $t_k \notin \{t_i: i\in\mathcal{I}_{-(v)}\}$. We then derive the validation score on the $v$th validation set: 
$$\text{CV}_v(t_k;\lambda_k,d_k,h)=\text{tr}\left(\hat{\bm{\Omega}}^\text{rf}_{-(v)}(t_k;d_k,\lambda_k,h)\hat{\bm{\Sigma}}_{(v)}(t_k)\right)-\log|\hat{\bm{\Omega}}^\text{rf}_{-(v)}(t_k;d_k,\lambda_k,h)|,$$
where $\hat{\bm{\Sigma}}_{(v)}(t_k):=\sum_{i\in \mathcal{I}_{(v)}}\omega_{h_V}^{t_i}(t_k)\bm{x}_i\bm{x}_i^T$ is the kernel estimate of the covariance matrix $\bm{\Sigma}(t_k)$  based on the $v$th validation set $\{\bm{x}_i\}_{i\in \mathcal{I}_{(v)}}$. Here, the bandwidth $h_V$ is set to be $h\cdot(\frac{1}{V-1})^{-1/5}$ to reflect the difference in sample sizes between the validation and training sets. Finally,  the $V$-fold cross-validation score at time $t_k$ is defined as: $\text{CV}(t_k;\lambda_k,d_k,h)=\sum_{v=1}^V\text{CV}_v(t_k;\lambda_k,d_k,h)$. The ``optimal" tuning parameters at $t_k$ given $h$, $(\hat{d}_k(h), \hat{\lambda}_k(h))$, is the pair that minimizes the CV score. Finally, the ``optimal" $h$ is chosen by minimizing the sum of  $\text{CV}(t_k;\hat{\lambda}_k(h),\hat{d}_k(h),h)$ over those time points where a \texttt{loggle} model is fitted. 

We also adopt the \texttt{cv.vote} procedure proposed in \cite{peng2010regularized} which has been shown to be able to significantly reduce the false discovery rate while  sacrifice only modestly in power. Specifically, given the CV selected tuning parameters, we examine the fitted  model on each training set and only retain those edges that appear in at least T\% of these models. In practice, we recommend 80\% as the cut off value for edge retention. 

Moreover, we implement efficient grid search strategies including early stopping and coarse search followed by refined search to further speed up the computation. Details can be found in Section \ref{sec:supp_tuning} of the Supplementary Material.

\section{Simulation}\label{sec:simulation}
In this section, we evaluate the performance of \texttt{loggle} and compare it with \texttt{kernel} and \texttt{invar} by simulation experiments.

\subsection{Setting}\label{sec:simulation_setting}
We consider models with both time-varying graphs and time-invariant graphs: 

\begin{itemize}
	\item
	\textit{Time-varying graphs}: (i) Generate four lower triangular matrices $\bm{B}_1,\;\bm{B}_2,\;\bm{B}_3,\;\bm{B}_4\in\mathbb{R}^{p\times p}$ with elements independently drawn from $\mathcal{N}(0,1/2)$. (ii) Let $\phi_1(t)=\sin(\pi t/2)$, $\phi_2(t)=\cos(\pi t/2)$, $\phi_3(t)=\sin(\pi t/4)$ and $\phi_4(t)=\cos(\pi t/4)$, $t\in[0,1]$, and set $\bm{G}(t)=\left(\bm{B}_1\phi_1(t)+\bm{B}_2\phi_2(t)+\bm{B}_3\phi_3(t)+\bm{B}_4\phi_4(t)\right)/2$. 
	(iii) Define $\bm{\Omega}^o(t)=\bm{G}(t)\bm{G}^T(t)$ and   ``soft threshold" its off-diagonal elements  to obtain $\bm{\Omega}(t)$: $\Omega_{uv}(t)=\text{sign}(1-\frac{0.28}{|\Omega^o_{uv}(t)|})\cdot(1-\frac{0.14}{|\Omega^o_{uv}(t)|})_+\Omega^o_{uv}(t)$, where $(x)_+=\max\{x, 0\}$. (iv) Add $\log_{10}(p)/4$ to the diagonal elements of $\bm{\Omega}(t)$ to ensure positive definiteness. 
	
	\item
	\textit{Time-invariant graphs}: (i) Generate an Erdos-Renyi graph \citep{erdos1959random} where each pair of nodes is connected independently with probability $2/p$ (so the total number of edges is around $p$). Denote  the edge set of this graph by $\mathbb{E}$. (ii) For off-diagonal elements ($1 \leq u \not= v \leq p$ ), if $\{u,v\}\notin \mathbb{E}$,  set $\Omega_{uv}(t) \equiv 0$ for  $t\in[0, 1]$;  If $\{u,v\}\in \mathbb{E}$, set $\Omega_{uv}(t)=\sin(2\pi t-c_{uv})$, where $c_{uv}\sim\text{uniform}(0, 1)$ is a random offset. (iii) For diagonal elements ($1\leq u  \leq p$), set $\Omega_{uu}(t)=|\sin(2\pi t-c_{uu})|+\log_{10}(p)$, where $c_{uu}\sim \text{uniform}(0, 1)$ is a random offset.
	
\end{itemize}

We construct three models following the above descriptions. Specifically, two models have time-varying graphs with $p=100$ and $p=500$ nodes, respectively.  In these two models,  the graphs change smoothly over time with the average number of edges being 51.6 (standard deviation 6.0) for $p=100$ model and 203.0 (standard deviation 66.8) for $p=500$ model. The plots depicting the number of edges vs. time  are given in Figure \ref{fig:ed_num}  of the Supplementary Material. In the third model, the graphs  are time-invariant (even though the precision matrices change over time)  with $p=100$ nodes and 93 fixed edges.

For each model, we  generate $\bm{x}_k\sim\mathcal{N}_p(\bm{0},\bm{\Omega}^{-1}(t_k))$, with $t_k=\frac{k-1}{N}$ ($k=1,\cdots, N+1$). 
We use the Epanechnikov kernel $K_h(x)=\frac{3}{4}\left(1-(x/h)^2\right)I_{\{|x|\leq h\}}$ to obtain smoothed estimates   of the correlation matrices. In the following, we consider $N=1000$  observations and conduct model fitting on $K=49$ time points at $\tilde{t}_k  \in \{0.02, 0.04, \ldots, 0.96, 0.98\}$.


We use 5-fold cross-validation for tuning parameters selection from $h_{\rm grid}=\{$0.1, 0.15, $\ldots$, 0.3$\}$, $d_{\rm grid}=\{$0, 0.001, 0.01, 0.025, 0.05, 0.075, 0.1, 0.15, 0.2, 0.25, 0.3, 1$\}$ and $\lambda_{\rm grid}=\{$0.15, 0.17, $\ldots$, 0.35$\}$. 

The metrics used for performance evaluation include false discovery rate:  $\text{FDR}:=1-\frac{1}{K}\sum_{k=1}^K|\hat{S}_k\cap S_k|/|\hat{S}_k|$ and 
$\text{power}:=\frac{1}{K}\sum_{k=1}^K|\hat{S}_k\cap S_k|/|S_k|$ for edge detection (averaged over the $K$ time points where graphs are estimated),  where $S_k$ and $\hat{S}_k$ are the true edge set and the estimated edge set at time point $\tilde{t}_k$, respectively. We also consider   $F_1:=2\cdot\frac{(1-\text{FDR})\cdot\text{power}}{(1-\text{FDR})+\text{power}}$ as an overall metric for  model selection performance which strikes a balance between FDR and power: The larger $F_1$ is, the better a method performs in terms of edge selection. In addition, we calculate the Kullback-Leibler (K-L) divergence (relative entropy) between the true models and the estimated models:  $\delta_{KL}:=\frac{1}{K}\sum_{k=1}^K[\text{tr}(\hat{\bm{\Omega}}(\tilde{t}_k)\bm{\Omega}^{-1}(\tilde{t}_k))-\log|\hat{\bm{\Omega}}(\tilde{t}_k)\bm{\Omega}^{-1}(\tilde{t}_k)|-p]$. 

\subsection{Results}\label{sec:eval_result}



\begin{table}[H]
	\begin{center}
		\caption{\textbf{Simulation Results.}  \label{table:model_selection}}
		\begin{tabular}{c|ccc|c}
			\hline\hline
			\multicolumn{5}{c}{$p=100$ time-varying graphs model}\\
			Method &  FDR & power & $F_1$&$\delta_{KL}$\\
			\hline
			\texttt{loggle} & 0.196 & 0.702 & 0.747 & 2.284\\
			\texttt{kernel}  & 0.063 & 0.571 & 0.703 & 2.690\\
			\texttt{invar} & 0.583 & 0.678 & 0.514 & 2.565\\
			\hline\hline
			
			\multicolumn{5}{c}{$p=500$ time-varying graphs model}\\
			Method & FDR & power & $F_1$&$\delta_{KL}$\\
			\hline
			\texttt{loggle} &  0.215 & 0.613 & 0.678 & 9.564\\
			\texttt{kernel} &  0.035 & 0.399 & 0.561 & 11.818\\
			\texttt{invar} &  0.590 & 0.597 & 0.478 & 10.608\\
			\hline\hline
			\multicolumn{5}{c}{$p=100$ time-invariant graphs model}\\
			Method & FDR & power & $F_1$&$\delta_{KL}$\\
			\hline
			\texttt{loggle} &  0.000 & 0.978 & 0.988 & 1.559\\
			\texttt{kernel} & 0.042 & 0.509 & 0.598 & 3.168\\
			\texttt{invar} & 0.000 & 1.000 & 1.000 & 1.531\\
			\hline\hline
		\end{tabular}
	\end{center}
\end{table}

Table \ref{table:model_selection} shows that under the time-varying graphs models,
\texttt{loggle}  outperforms \texttt{kernel} according to $F_1$ score and K-L divergence. 
Not surprisingly, \texttt{invar} performs very poorly for time-varying graphs models. On the other hand, under the time-invariant graphs model, \texttt{loggle} performs similarly as \texttt{invar}, whereas \texttt{kernel} performs very poorly.
These results demonstrate that \texttt{loggle} can adapt to different degrees of smoothness of the graph topology in a data driven fashion and has generally good performance across a wide range of scenarios including both time-varying and time-invariant graphs.

The \texttt{loggle} procedure is more computationally intensive than \texttt{kernel} and \texttt{invar} as it fits many more models. 
For the $p=100$  time-varying graphs model, \texttt{loggle}  took 3750 seconds using 25 cores on a linux server with 72 cores, 256GB RAM and two Intel Xeon E5-2699 v3 @ 2.30GHz processors. At the same time, \texttt{kernel} took  226 seconds and \texttt{invar} took 777 seconds. 
 On average, per \texttt{loggle}  model fit took 23.2 milliseconds (ms), per \texttt{kernel} model fit took 16.8ms and  per \texttt{invar} model fit took 2825.5 ms.  The additional computational cost of \texttt{loggle}  is justified by its superior performance and should become less of a burden with fast growth of computational power.

\section{S\&P 500 Stock Price} \label{sec:real_data}

In this section, we apply \texttt{loggle} to the S\&P 500 stock price dataset obtained via R package \texttt{quantmod} from \textit{www.yahoo.com}. We focus on 283 stocks from 5 Global Industry Classification Standard (GICS) sectors: 58 stocks from  Information Technology, 72 stocks from Consumer Discretionary, 32 stocks from Consumer Staples, 59 stocks from Financials, and 62 stocks from Industrials. We are interested in elucidating how direct interactions (characterized by conditional dependencies) among these stocks are evolving over time and particularly how such interactions are affected by the recent global financial crisis.

For this purpose, we consider a 4-year  time period from  January 1st, 2007 to January 1st, 2011, which covers the recent global financial crisis: ``According to the U.S. National Bureau of Economic Research, the recession, as experienced in that country, began in December 2007 and ended in June 2009, thus extending over 19 months. The Great Recession was related to the financial crisis of 2007-2008 and U.S. subprime mortgage crisis of 2007-2009 (Source: wikipedia)". Each stock has 1008 closing prices during this period, denoted by $\{y_k\}_{k=1}^{1008}$.  We use the logarithm of the ratio between two adjacent prices, i.e., $\log \frac{y_{k+1}}{y_{k}} (k=1,\cdots, 1007)$ for the subsequent analysis. We also convert the time points onto  [0, 1] by $t_k=\frac{k-1}{1006}$ for $k=1,\cdots, 1007$.  By examining the autocorrelation (Figure \ref{fig:acf} of the Supplementary Material), the independence  assumption appears to hold reasonably well. 

We use the Epanechnikov kernel to obtain the kernel estimates of the correlation matrices. 
We then fit  three models, namely, \texttt{loggle}, \texttt{kernel} and \texttt{invar},  at $K=201$ time points $\{0.005, 0.010, \ldots, 0.995\}$ using 5-fold cross-validation for model tuning.  We use the tuning grids $h_{\rm grid}=\{0.1, 0.15\}$, $\lambda_{\rm grid}=\{10^{-2}, 10^{-1.9}, \ldots, 10^{-0.1}, 1\}$ and 
$d_{\rm grid}=\{0$, 0.001, 0.01, 0.025, 0.05, 0.075, 0.1, 0.15, 0.2, 0.25, 0.3, 1$\}$, where $h_{\rm grid}$ is pre-selected by using coarse search described in Section \ref{sec:supp_tuning} of the Supplementary Material.  Table \ref{table:model_selection_real} reports  the average number of edges across the fitted  graphs (and standard deviations in parenthesis) as well as the CV scores. We can see that \texttt{loggle} has a significantly smaller CV score than those of  \texttt{kernel} and \texttt{invar}. Moreover, on average, \texttt{loggle} and \texttt{invar} models have similar number of edges, whereas \texttt{kernel} models have more edges. 
\begin{table}[h]
	\begin{center}
		\caption{\textbf{Stock price:} Number of edges and CV score}\label{table:model_selection_real}
		
		\begin{tabular}{c|cc}
			\hline\hline
			Method & Average edge \#  (s.d.) & CV score\\
			\hline
			\texttt{loggle} & 819.4 (331.0) & 123.06\\
			\texttt{kernel} &  1103.5 (487.1) & 160.14\\
			\texttt{invar} & 811.0 (0.0) &130.68\\
			\hline\hline
		\end{tabular}
	\end{center}
\end{table}

\begin{figure}[h]
	\begin{center}
		\begin{tabular}{cc}
			\includegraphics[width = 2.8in]{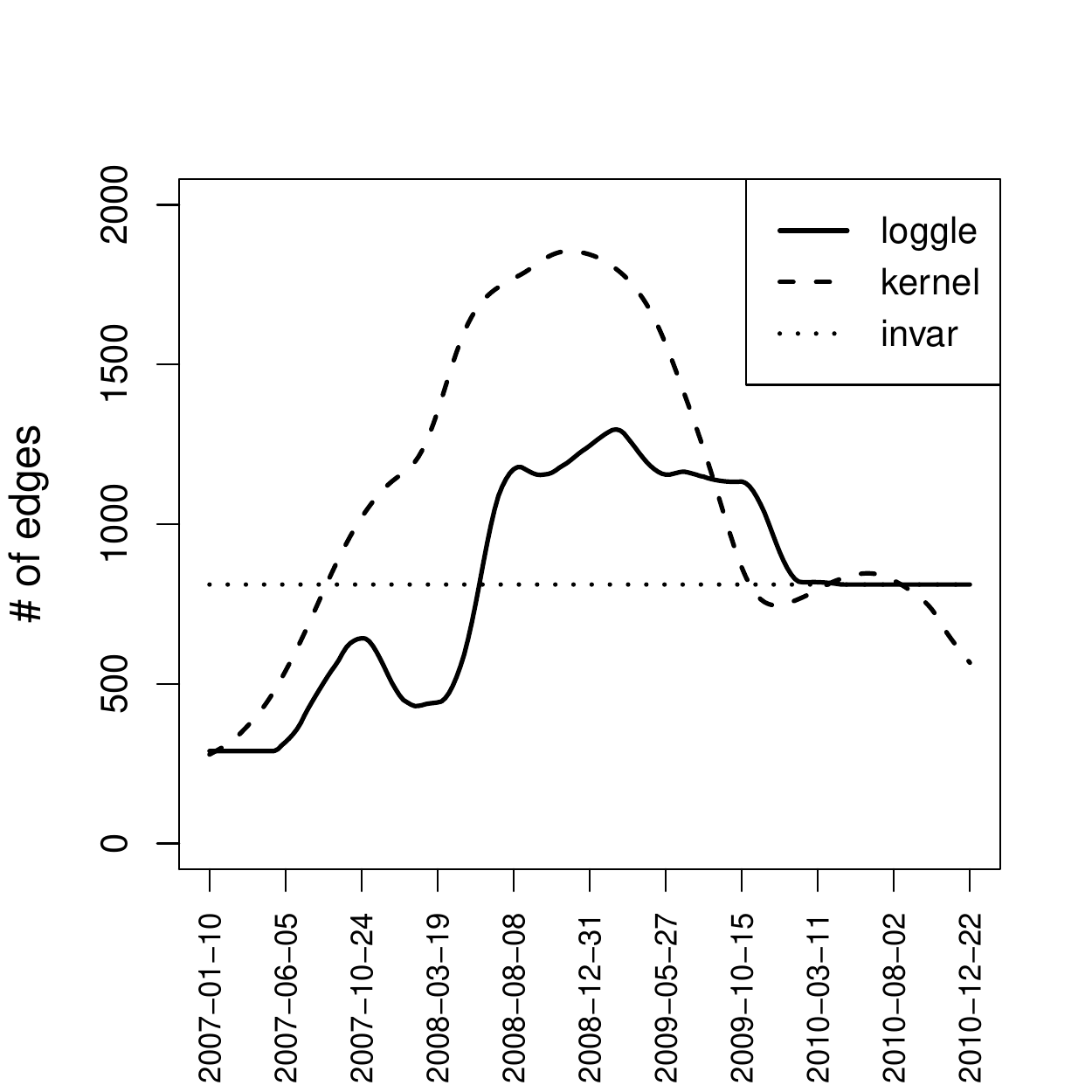} & \includegraphics[width = 2.8in]{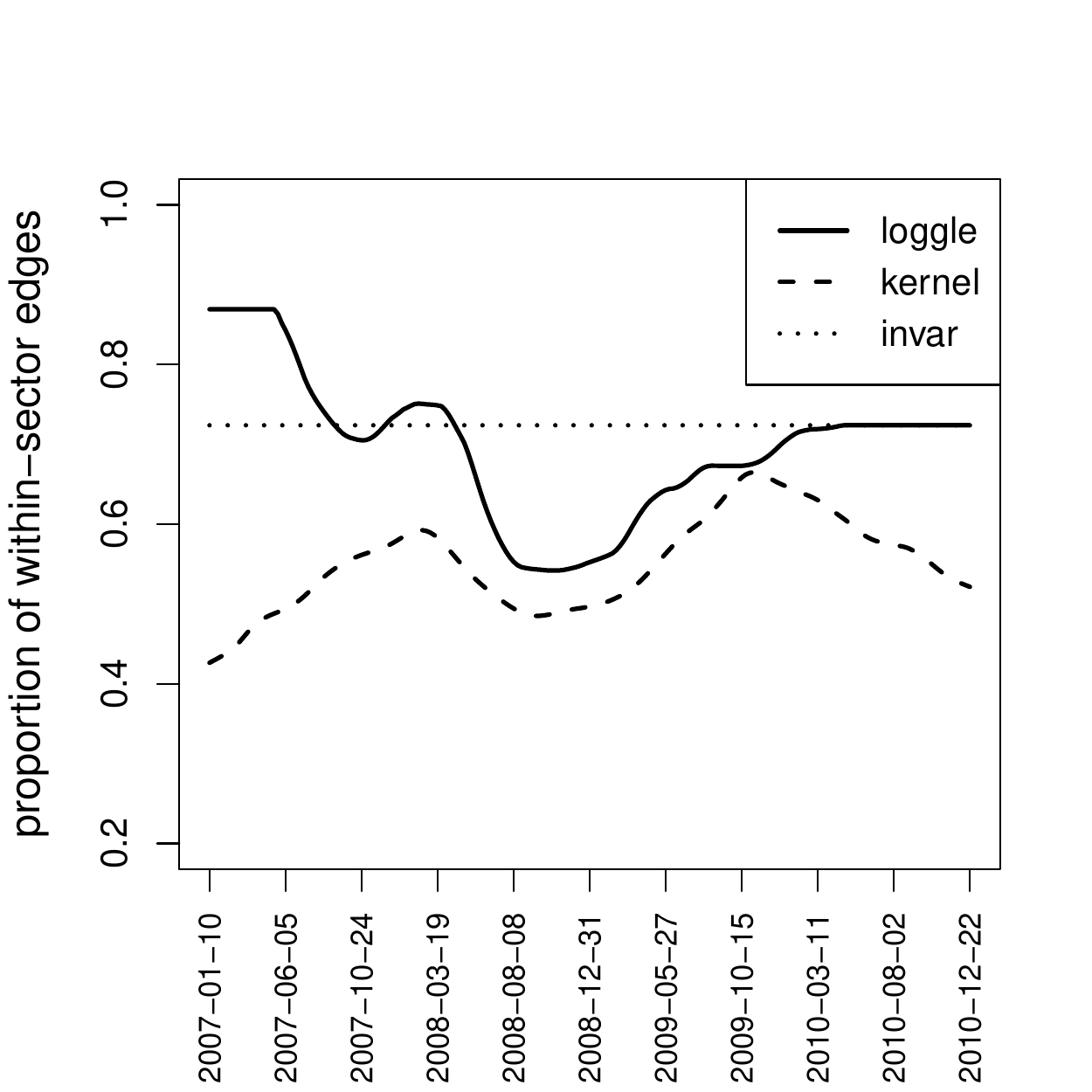}\\
			(a) & (b)
		\end{tabular}
	\end{center}
	\caption{\textbf{Stock Price.} (a) Number of edges vs. time  (b) Proportion of within-sector edges vs. time}\label{fig:ed}
\end{figure}


Figure \ref{fig:ed}(a) shows the number of edges in the fitted graph over time.  
The \texttt{invar} fitted graphs have an identical topology, which is unable to reflect the evolving relationships among these stocks. On the other hand,  both \texttt{loggle} and \texttt{kernel} are able to capture the changing relationships by fitting graphs with time-varying topologies. More specifically, both methods detect an increased amount of interaction (characterized by larger number of edges) during the financial crisis. The amount of interaction  peaked around the end of 2008 and then went down to a level still higher than that of the pre-crisis period. As can be seen from the figure, the \texttt{kernel} graphs show rather drastic changes, whereas the \texttt{loggle} graphs change more gradually.  The \texttt{loggle} method in addition detects a period with increased interaction in the early stage of the financial crisis, indicated by the  smaller peak around October 2007 in Figure  \ref{fig:ed}(a). This is likely due to the  subprime mortgage crisis which  acted as a precursor of the   financial crisis \citep{amadeo_2017_07}. In the period after the financial crisis, the \texttt{loggle} fits are similar to those of \texttt{invar} with a nearly constant  graph topology after March 2010, indicating that the relationships among the stocks had stabilized. In contrast, \texttt{kernel} fits show a small bump in edge number around the middle of 2010 and decreasing amount of interaction  afterwards.

Figure \ref{fig:ed}(b) displays the proportion of  within-sector edges among the total number of detected edges. During the entire time period, \texttt{loggle} fitted graphs consistently have higher proportion of within-sector edges  than that of the \texttt{kernel} fitted graph. For both methods,  this proportion decreased during the financial crisis due to increased amount of cross-sector interaction. For \texttt{loggle}, the within-sector edge proportion eventually increased and stabilized after March 2010, although at a level lower than that of the pre-crisis period. In contrast, for \texttt{kernel}, the within-sector proportion  took a downturn again after October 2009. 
In summary, the \texttt{loggle} fitted graphs are easier to interpret in terms of describing the evolving interacting relationships among the stocks and  identifying the underlying sector structure of the stocks. Hereafter, we focus our discussion on \texttt{loggle} fitted graphs. 


\begin{figure}[h]
	\begin{center}
		\begin{tabular}{ccc}
			\includegraphics[width = 2.1in]{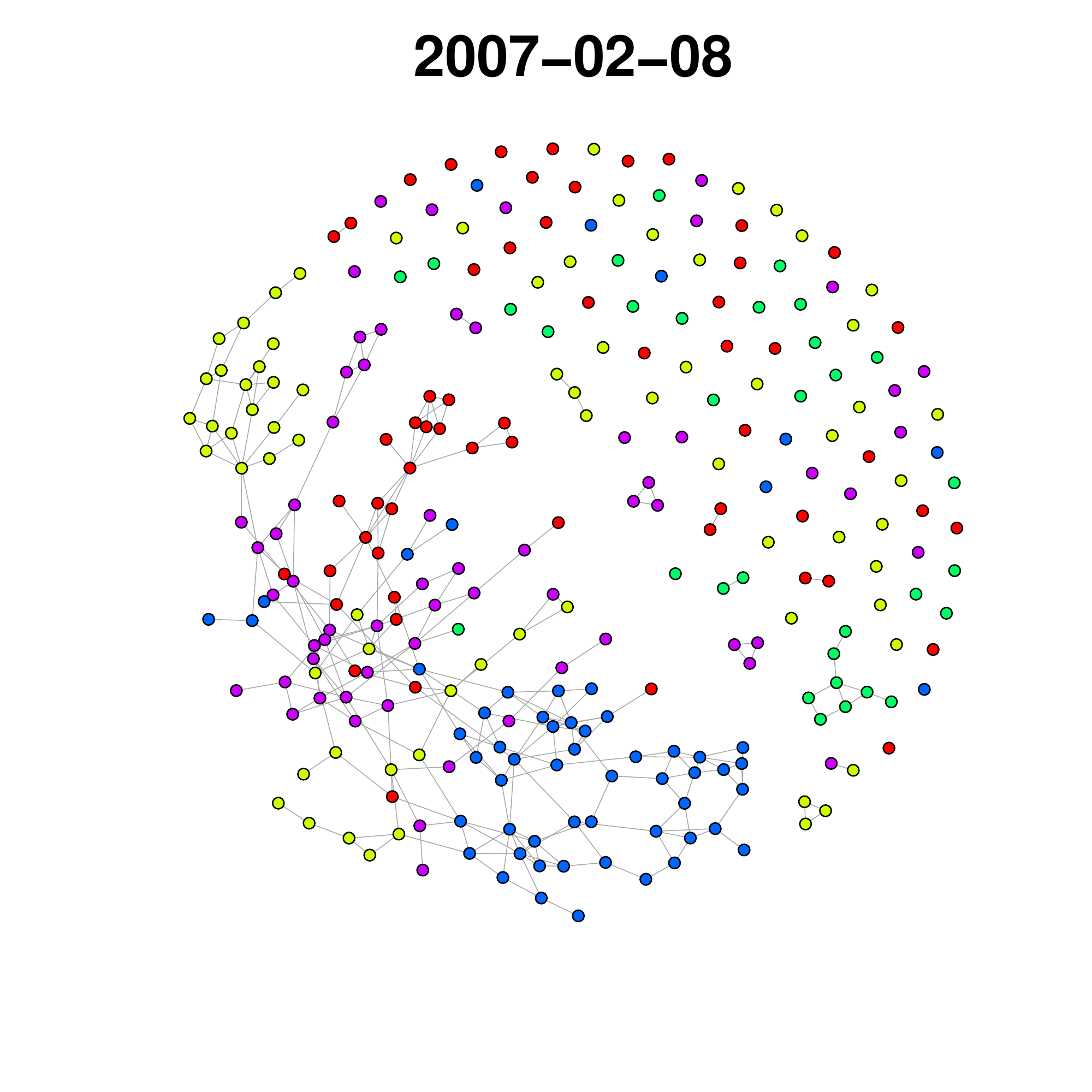} & \includegraphics[width = 2.1in]{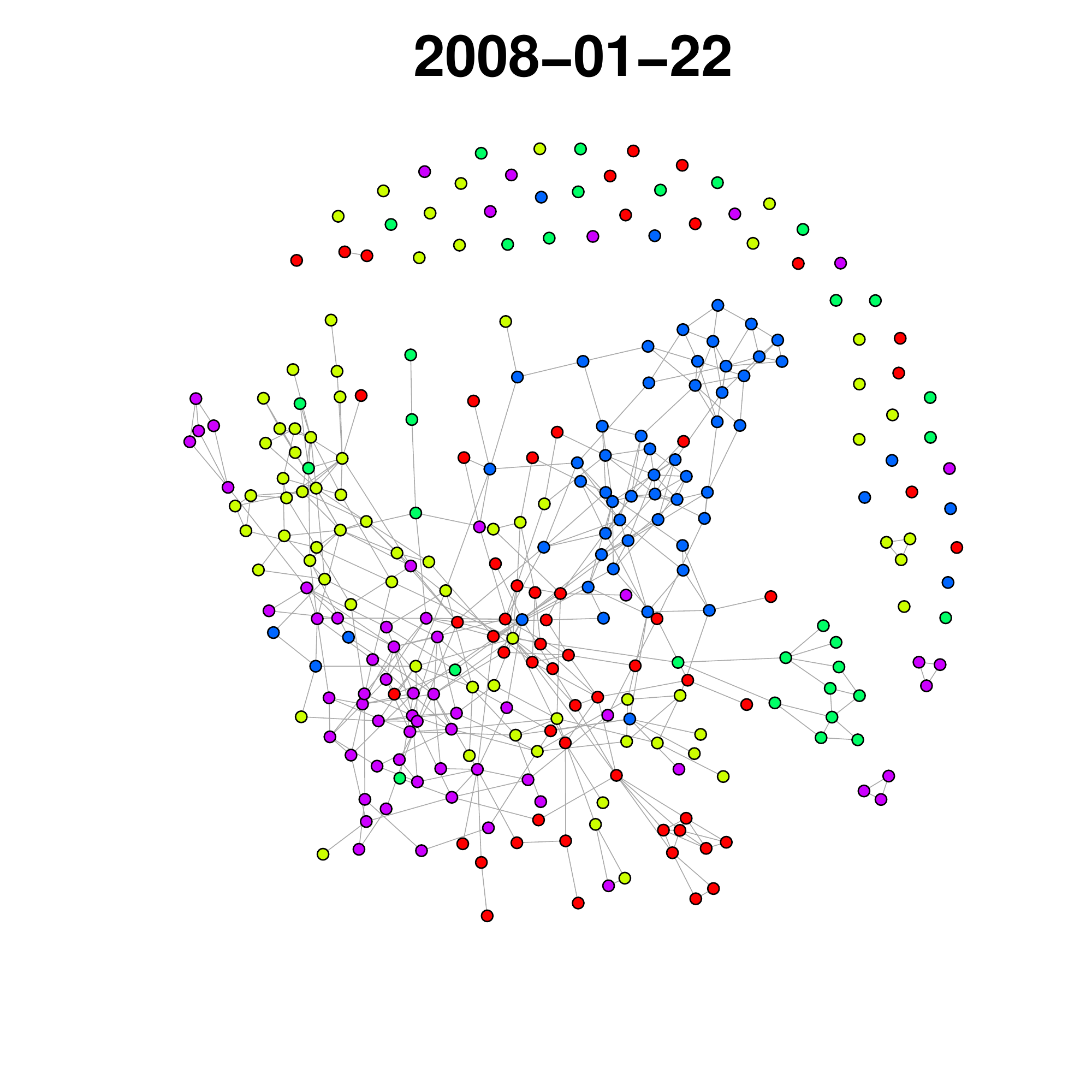} & \includegraphics[width = 2.1in]{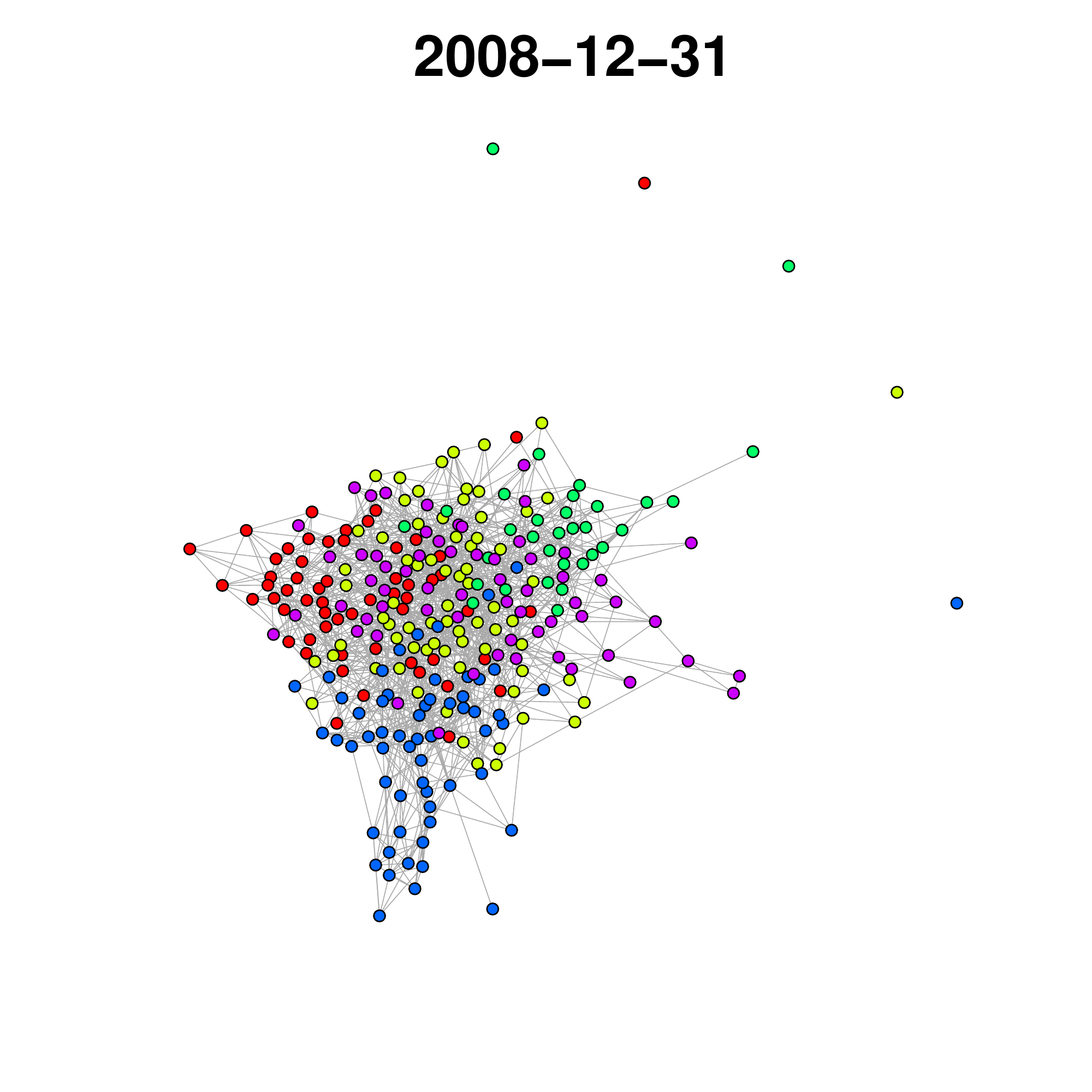}\\
			(a) & (b) & (c)\\
			\includegraphics[width = 2.1in]{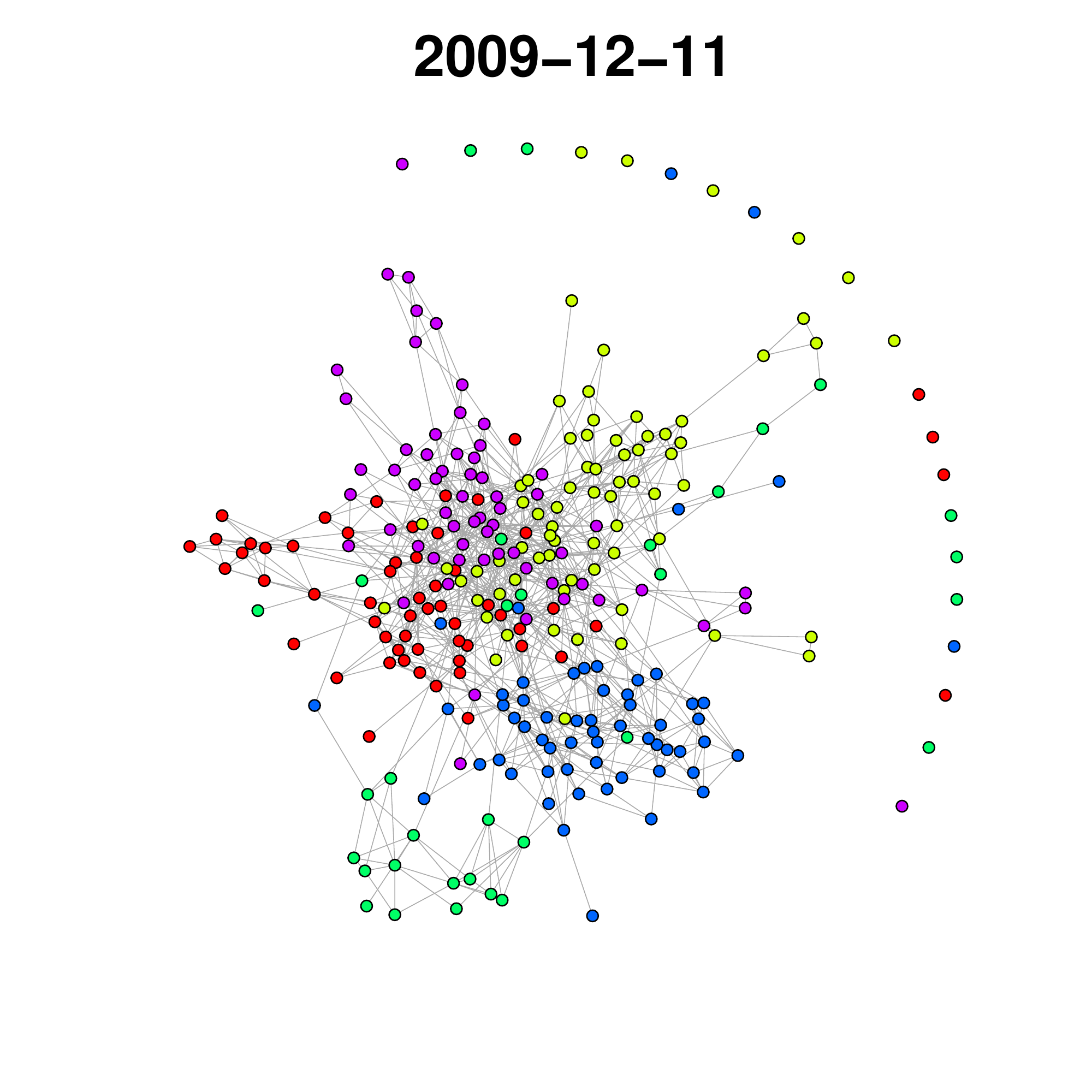} & \includegraphics[width = 2.1in]{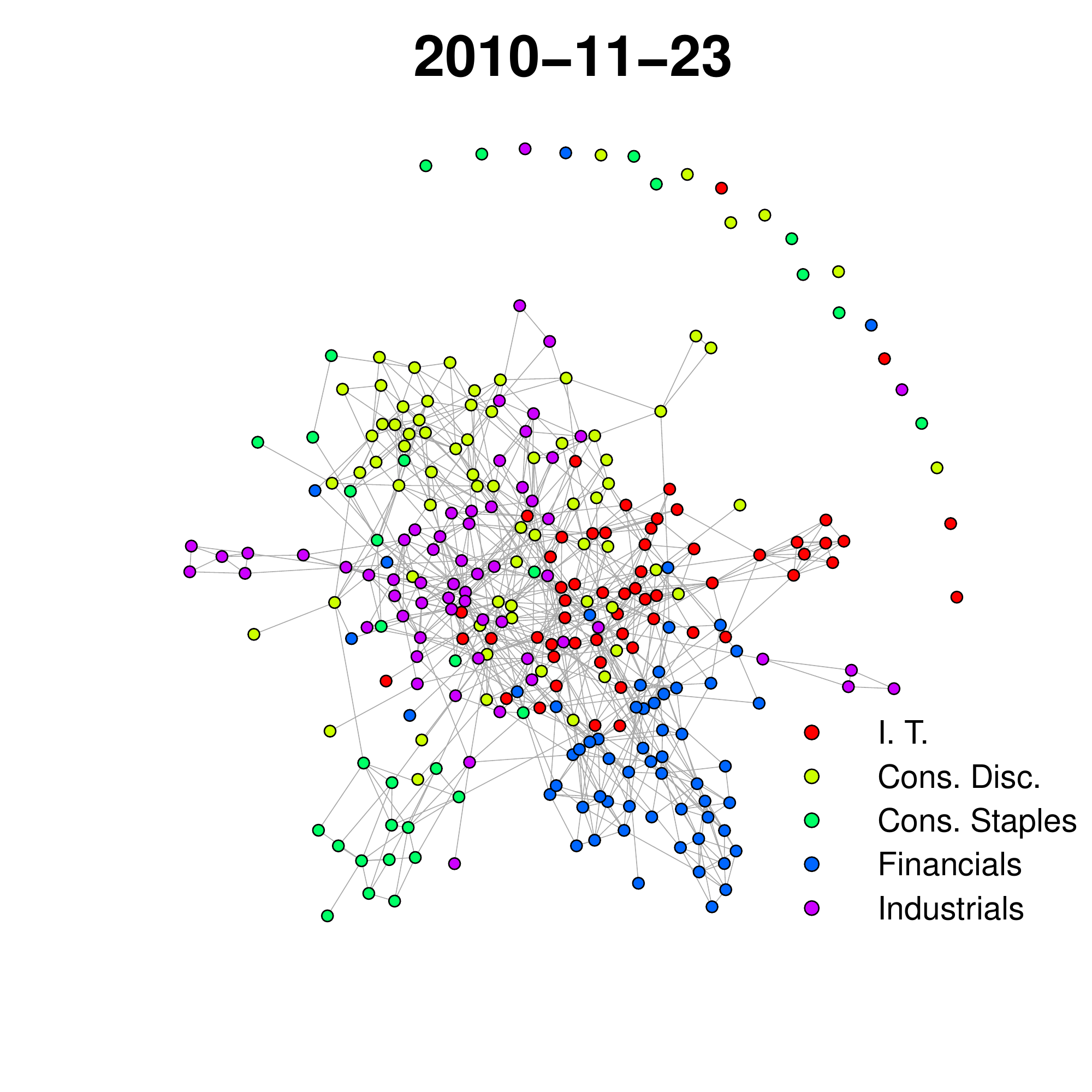} & \includegraphics[width = 2.1in]{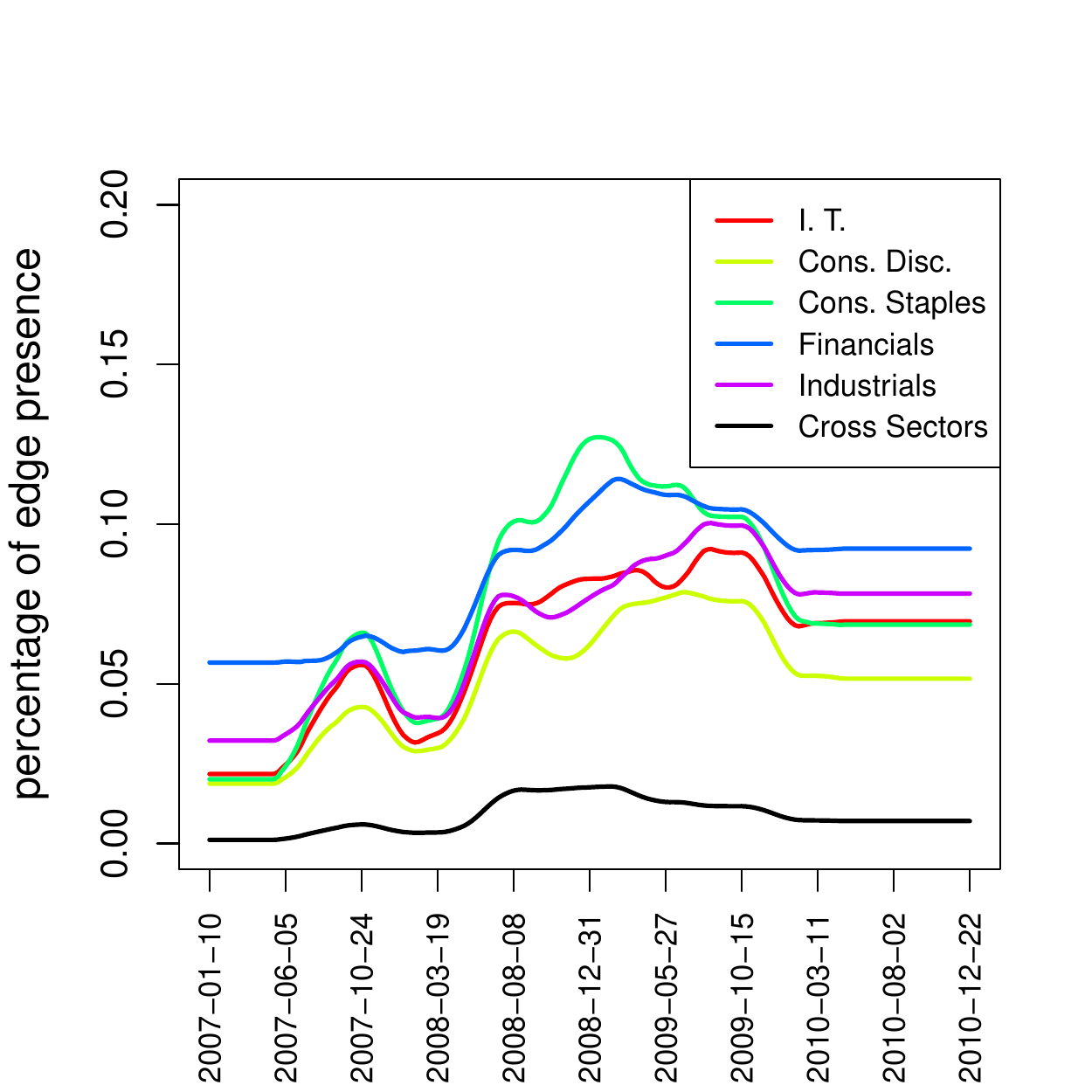}\\
			(d) & (e) & (f)
		\end{tabular}
	\end{center}
	\caption{\textbf{Stock Price.}  (a)-(e) \texttt{loggle}  fitted graphs at 5 time points. (f) The sector-wise percentage of presence of within-sector edges   and the percentage of presence of cross-sector edges  of  \texttt{loggle} fitted graphs vs. time.}\label{fig:graph_loggle}
\end{figure}

Figure \ref{fig:graph_loggle}(a)-(e) show the \texttt{loggle} fitted graphs  at 5 different time points, namely, before, at the early stage, around the peak, towards the end and after the financial crisis. These graphs show clear evolving interacting patterns among the stocks. The amount of interaction increased with the deepened crisis and decreased and eventually stabilized with the passing of the crisis. Moreover, the stocks have more interactions  after the   crisis compared to the pre-crisis era, indicating fundamental change of the financial landscape. 
In addition, these graphs show clear sector-wise clusters (nodes with the same color corresponding to stocks from the same sector).

Figure \ref{fig:graph_loggle}(f) shows the sector-wise \textit{percentage of presence of within-sector edges}, defined as the ratio between the number of detected within-sector edges and the total number of possible within-sector edges for a given sector; and  the \textit{percentage of  presence of cross-sector edges}, defined as the ratio between the number of detected cross-sector edges and the total number of possible cross-sector edges.  As can be seen from this figure,  the within-sector percentages are much higher than the cross-sector percentage, reaffirming the observation that \texttt{loggle} is able to identify the underlying sector structure.  Moreover, the within-Financials sector  percentage is among the highest across  the entire time period, indicating that the stocks in this sector have been consistently highly interacting with each other.   Finally, all percentages increased after the financial crisis began and leveled off afterwards, reflecting the increased amount of interaction during the financial crisis. 

\begin{figure}[h]
	\begin{center}
		\begin{tabular}{ccc}
			\includegraphics[width = 2.1in]{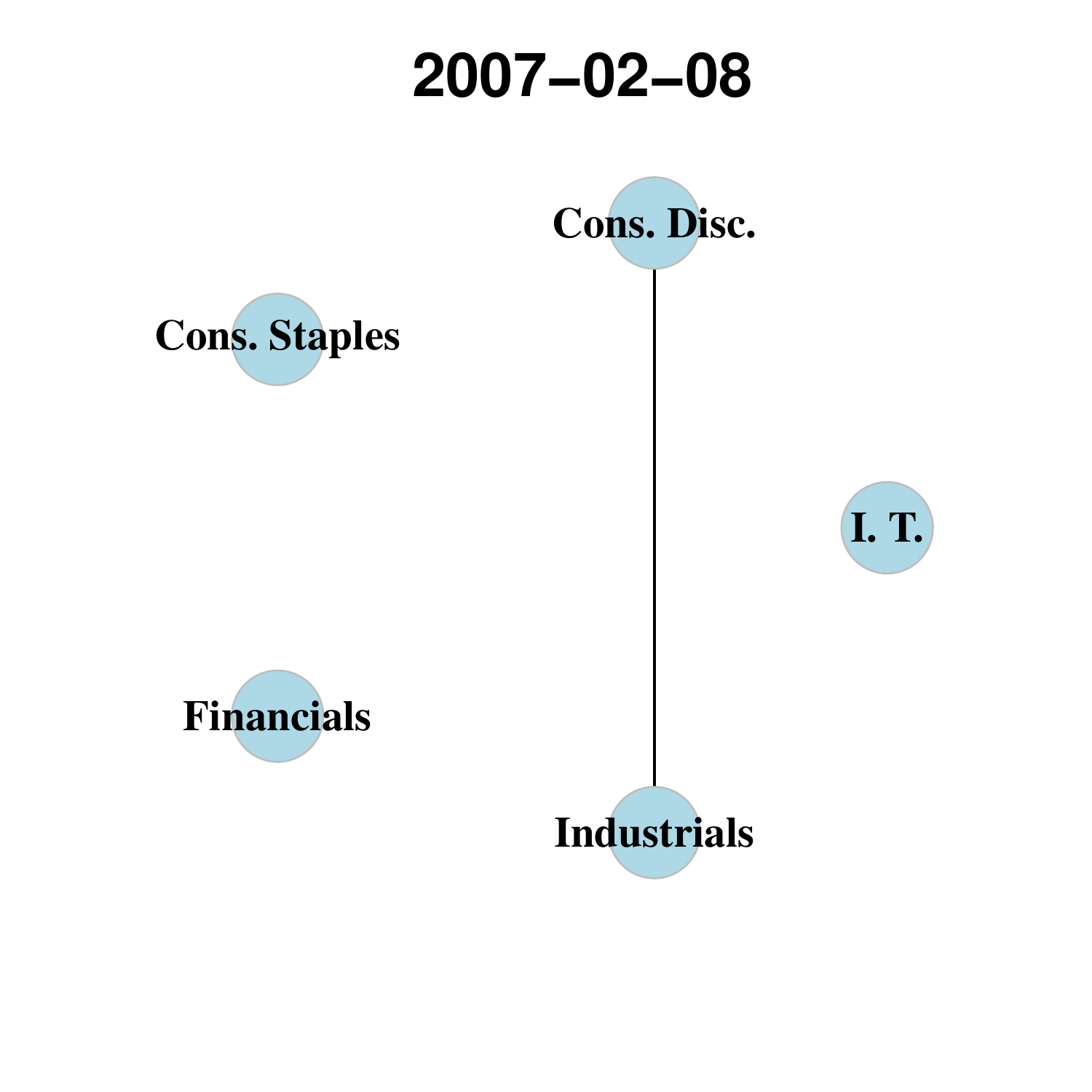} & \includegraphics[width = 2.1in]{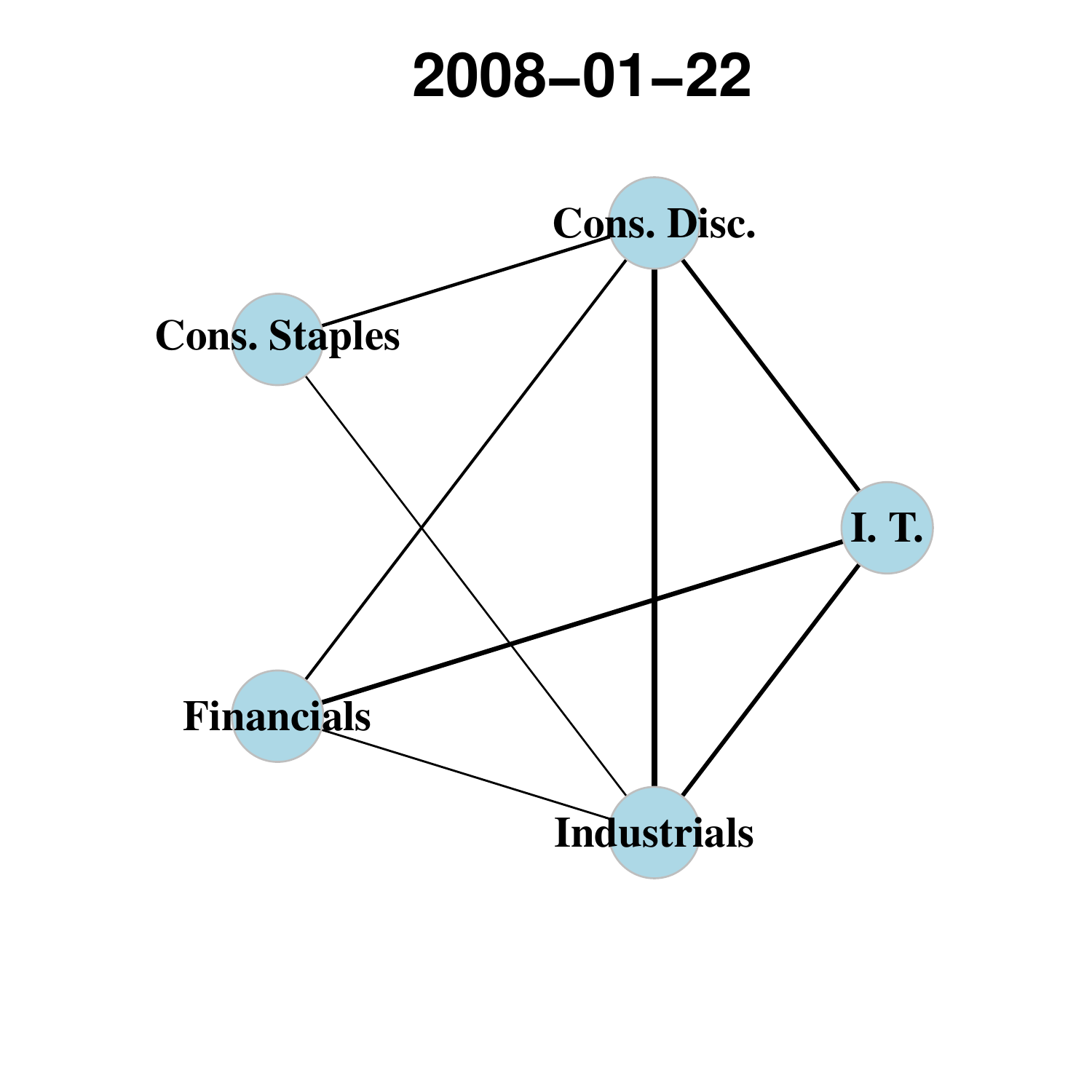} & \includegraphics[width = 2.1in]{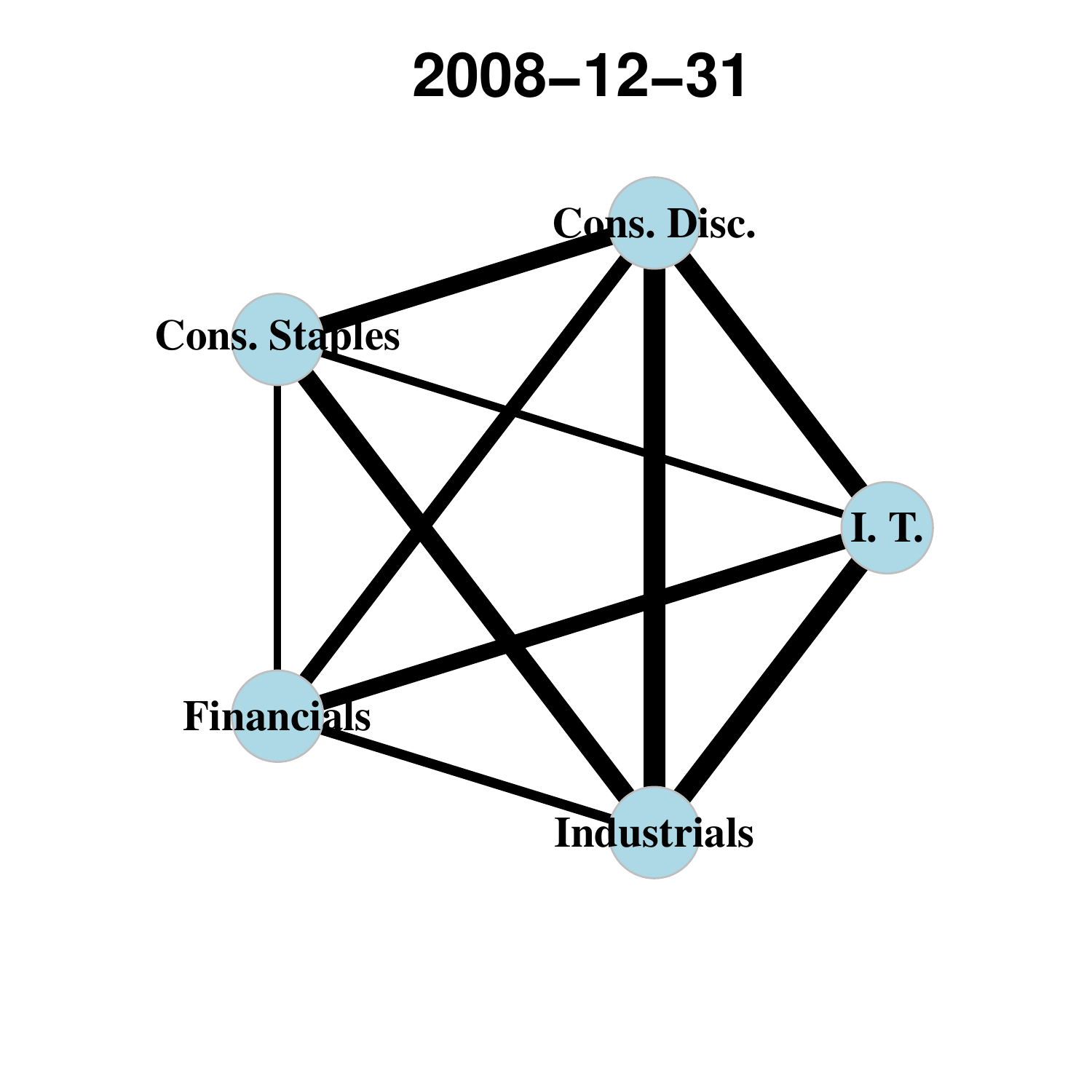}\\
			\includegraphics[width = 2.1in]{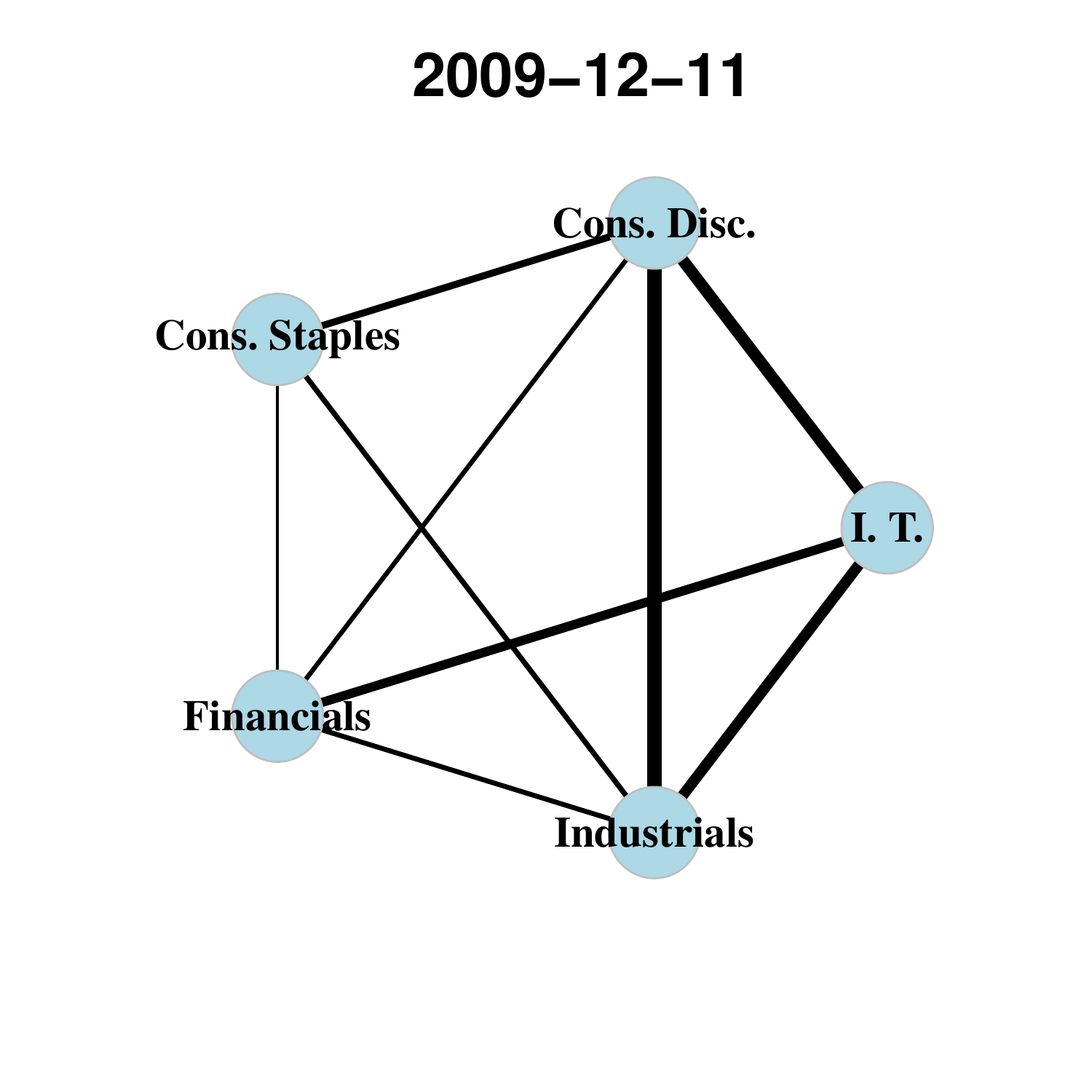} & \includegraphics[width = 2.1in]{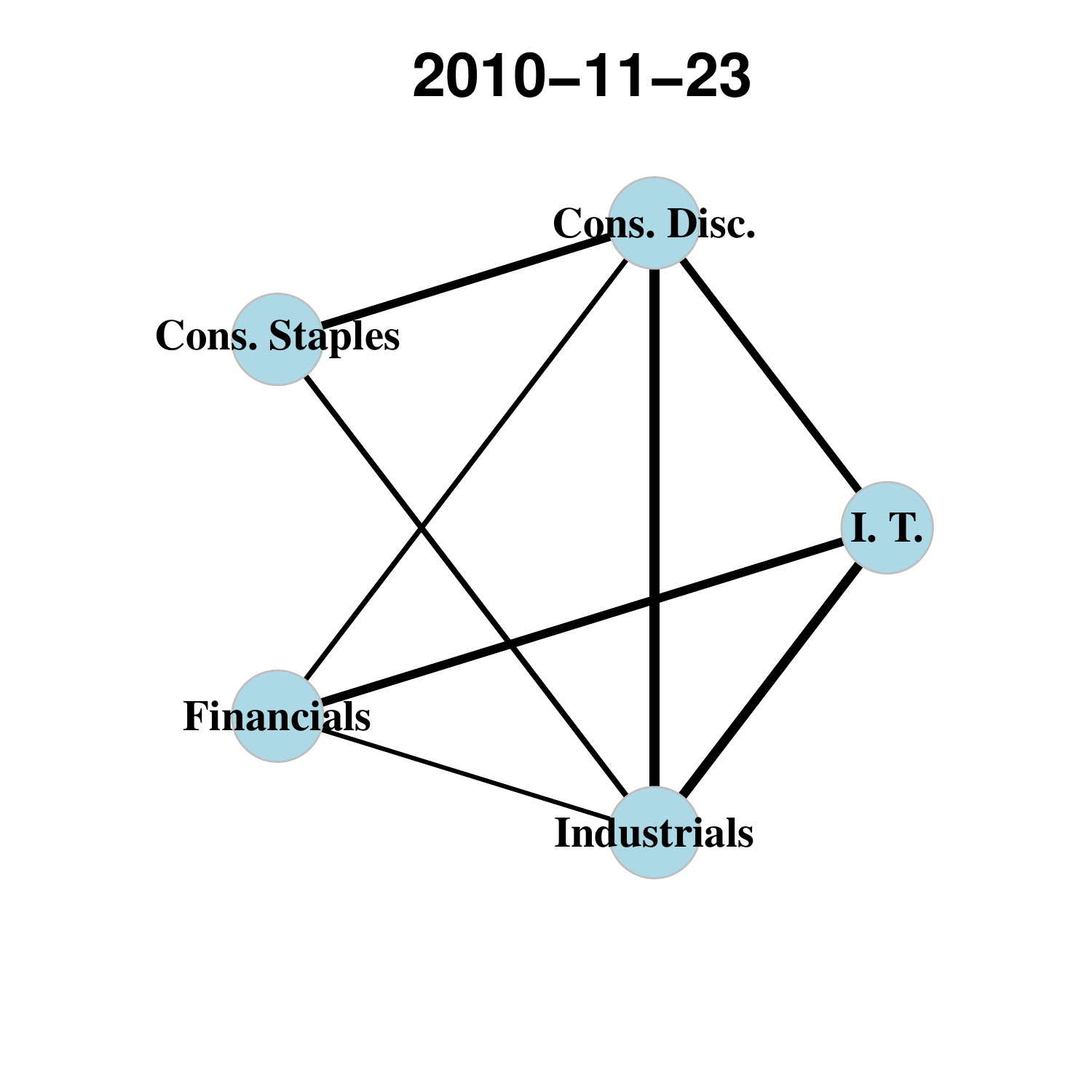} & \
		\end{tabular}
	\end{center}
	\caption{\textbf{Stock Price.}  Cross-sector interaction plots at 5 different time points based on \texttt{loggle} fitted graphs. Each node represents a sector and the edge width is proportional to the percentage of presence of the corresponding cross-sector edges.}\label{fig:intersector_loggle}
\end{figure}

In Figure \ref{fig:intersector_loggle}, the graphs describe cross-sector  interactions among the 5 GICS sectors at five different time points (before, at early stage, around the peak, at late stage and after the financial crisis). In these graphs, each node represents a sector and edge width is proportional to the respective  percentage of presence of cross-sector edges (defined as the detected number of edges between two sectors divided by the total number of possible  edges between these two sectors). Moreover, edges with cross-sector percentage less than $0.2\%$ are not displayed. We can see that there are more cross-sector interactions during the financial crisis, indicating higher degree of dependency among different sectors in that period. There are also some interesting observations with regard to how these sectors interact with one another and how such interactions change over time. 
For example, strong cross-sector interactions between the Financials sector and the Consumer Staples sector arose during the financial crisis despite of their weak relationship before and after the crisis. This is probably due to  strong influence of the financial industry on the entire economy during financial crisis. Take the Consumer Discretionary sector and the Industrials sector as another example. These two sectors maintained a persistent relationship throughout the four years, indicating intrinsic  connections between  them irrespective of the financial landscape. 

\section{Conclusion}\label{sec:discussion}

In this paper, we propose \textit{LOcal Group Graphical Lasso Estimation} -- \texttt{loggle}, a novel model for estimating a sequence of time-varying graphs based on temporal observations. By using a local group-lasso type penalty, \texttt{loggle} imposes structural smoothness on the estimated graphs and consequently leads to more efficient use of the data as well as  more interpretable graphs. Moreover, \texttt{loggle} can  adapt to the local degrees of smoothness and sparsity of the underlying graphs in a data driven fashion and thus is effective under a wide range of scenarios. 
We develop a computationally efficient algorithm  for  \texttt{loggle}  that utilizes  the block-diagonal structure and pseudo-likelihood approximation. 
The effectiveness of \texttt{loggle}  is demonstrated through simulation experiments. Moreover, by applying \texttt{loggle} to the S\&P 500 stock price data, we  obtain interpretable and insightful graphs about the dynamic interacting relationships among the stocks, particularly on how such relationships change in response to the recent global financial crisis.
An R package \texttt{loggle} has been developed and  will be available on  \url{http://cran.r-project.org/}.

\section*{Appendix}
\setcounter{subsection}{0}
\renewcommand{\thesubsection}{A.\arabic{subsection}}
\setcounter{equation}{0}
\renewcommand{\theequation}{A.\arabic{equation}}

\subsection{Proof of Theorem \ref{thm:disconnect}}\label{sec:disconnect}

By the KKT conditions, a necessary and sufficient set of conditions for $\hat{\mathbb{\Omega}}_k=\{\hat{\bm{\Omega}}(t_i)\}_{i \in \mathcal{N}_{k,d}}$ being the minimizer of $L(\mathbb{\Omega}_k)$ in (\ref{eq:loggle}) is:
\begin{equation}\label{eq:disconnect}
\frac{1}{\sqrt{|\mathcal{N}_{k,d}|}}\left(\hat{\bm{\Sigma}}(t_i)-\hat{\bm{\Omega}}(t_i)^{-1}\right)+\lambda\bm{\Gamma}(t_i)=\bm{0},\ \forall i\in \mathcal{N}_{k,d},
\end{equation}
where $\bm{\Gamma}(t_i)=(\Gamma_{uv}(t_i))_{p\times p}$, and $(\Gamma_{uv}(t_i))_{i\in \mathcal{N}_{k,d}}$ is a subgradient of $\sqrt{\sum_{i\in \mathcal{N}_{k,d}}\Omega_{uv}(t_i)^2}$:
\begin{equation*}
(\Gamma_{uv}(t_i))_{i\in \mathcal{N}_{k,d}}
\begin{cases}
=\left(\frac{\Omega_{uv}(t_i)}{\sqrt{\sum_{i\in \mathcal{N}_{k,d}}\Omega_{uv}(t_i)^2}}\right)_{i\in \mathcal{N}_{k,d}} & \text{if}\ \sum_{i\in \mathcal{N}_{k,d}}\Omega_{uv}(t_i)^2>0\\
\text{such that}\ \sum_{i\in \mathcal{N}_{k,d}}\Gamma_{uv}(t_i)^2\leq1 & \text{if}\ \sum_{i\in \mathcal{N}_{k,d}}\Omega_{uv}(t_i)^2=0.
\end{cases}
\end{equation*}
If for $\forall i  \in  \mathcal{N}_{k,d}$, $\hat{\bm{\Omega}}(t_i)=\left(\begin{array}{c c}\hat{\bm{\Omega}}_1(t_i) & \bm{0} \\ \bm{0} &\hat{\bm{\Omega}}_2(t_i)\end{array}\right)$, where $\hat{\bm{\Omega}}_1(t_i)$ and $\hat{\bm{\Omega}}_2(t_i)$ consist of the variables in $G_1$ and $G_2$ respectively, then $\hat{\mathbb{\Omega}}_k=\{\hat{\bm{\Omega}}(t_i)\}_{i\in \mathcal{N}_{k,d}}$ satisfies  (\ref{eq:disconnect}) iff $\forall u\in G_1$, $\forall v\in G_2$, $\exists$ $(\Gamma_{uv}(t_i))_{i\in \mathcal{N}_{k,d}}$ satisfying $\sum_{i\in \mathcal{N}_{k,d}}\Gamma_{uv}(t_i)^2\leq1$ such that $$\frac{1}{\sqrt{|\mathcal{N}_{k,d}|}}\hat{\Sigma}_{uv}(t_i)+\lambda\Gamma_{uv}(t_i)=0,\ \forall i\in \mathcal{N}_{k,d}.$$
This is equivalent to
$$\forall u\in G_1,\ \forall v\in G_2,\ \frac{1}{|\mathcal{N}_{k,d}|}\sum_{i\in \mathcal{N}_{k,d}}\hat{\Sigma}_{uv}(t_i)^2\leq\lambda^2.$$

\subsection{ADMM algorithm under pseudo-likelihood approximation}\label{sec:ADMM_PL}

To solve the optimization problem in (\ref{eq:pseudo_concat}) using ADMM algorithm, we notice that the problem can be written as
\begin{align*}
&\textrm{minimize }_{\mathbb{B}_k,\mathbb{Z}_k}\sum_{i\in \mathcal{N}_{k,d}}\frac{1}{2}||\mathbf{Y}(t_i)-\mathbf{X}(t_i)\bm{\beta}(t_i)||_2^2
+\lambda\sum_{u<v}\sqrt{\sum_{i\in \mathcal{N}_{k,d}}\left[Z_{uv}(t_i)^2+Z_{vu}(t_i)^2\right]},\\
&\textrm{subject to }\bm{\beta}(t_i)-\bm{Z}(t_i)=\bm{0},\;i\in \mathcal{N}_{k,d},
\end{align*}
where $\mathcal{N}_{k,d}=\{i:|t_i-t_k|\leq d\}$, $\mathbb{B}_k=\{\bm{\beta}(t_i)\}_{i\in \mathcal{N}_{k,d}}$ and $\mathbb{Z}_k=\{\bm{Z}(t_i)\}_{i\in \mathcal{N}_{k,d}}$. Note $\bm{\beta}(t_i)=(\beta_{uv}(t_i))_{u\neq v}$ and  $\bm{Z}(t_i)=(Z_{uv}(t_i))_{u\neq v}$  are $\mathbb{R}^{p(p-1)}$ vectors.

The scaled augmented Lagrangian is
\begin{align*}
L_\rho(\mathbb{B}_k,\mathbb{Z}_k,\mathbb{U}_k)=&\sum_{i\in \mathcal{N}_{k,d}}\frac{1}{2}||\mathbf{Y}(t_i)-\mathbf{X}(t_i)\bm{\beta}(t_i)||_2^2
+\lambda\sum_{u<v}\sqrt{\sum_{i\in \mathcal{N}_{k,d}}\left[Z_{uv}(t_i)^2+Z_{vu}(t_i)^2\right]}\\
&+\frac{\rho}{2}\sum_{i\in \mathcal{N}_{k,d}}||\bm{\beta}(t_i)-\bm{Z}(t_i)+\bm{U}(t_i)||_2^2,
\end{align*}
where $\mathbb{U}_k=\{\bm{U}(t_i)\}_{i\in \mathcal{N}_{k,d}}$  are dual variables ($\bm{U}(t_i)=(U_{uv}(t_i))_{u\neq v}\in\mathbb{R}^{p(p-1)}$).

The ADMM algorithm is as follows. We first initialize $\bm{Z}^{(0)}(t_i)=\bm{0}$, $\bm{U}^{(0)}(t_i)=\bm{0}$, $i\in \mathcal{N}_{k,d}$. We also need to specify $\rho (>0)$, which in practice is recommended to be  $\approx\lambda$ \citep{wahlberg2012admm}. For step $s=1,2,\ldots$ until convergence:
\begin{enumerate}[(i)]
	\item
	For $i\in \mathcal{N}_{k,d}$,
	\begin{align*}
	\bm{\beta}^{s}(t_i)=\arg\min_{\bm{\beta}(t_i)}\frac{1}{2}||\mathbf{Y}(t_i)-\mathbf{X}(t_i)\bm{\beta}(t_i)||_2^2+\frac{\rho}{2}||\bm{\beta}(t_i)-\bm{Z}^{s-1}(t_i)+\bm{U}^{s-1}(t_i)||_2^2.
	\end{align*}
	The solution $\bm{\beta}^{s}(t_i)$ sets the derivative of the objective function to 0: 
	$$
	(\mathbf{X}(t_i)^T\mathbf{X}(t_i)+\rho\bm{I})\bm{\beta}^s(t_i)=\mathbf{X}(t_i)^T\mathbf{Y}(t_i)+\rho(\bm{Z}^{s-1}(t_i)-\bm{U}^{s-1}(t_i)).
	$$
	It is easy to see that
	$$
	\mathbf{X}(t_i)^T\mathbf{X}(t_i)+\rho\bm{I}={\rm diag}\{\tilde{\mathbf{X}}_{(-u)}(t_i)^T\tilde{\mathbf{X}}_{(-u)}(t_i)+\rho\bm{I}\}_{1\leq u\leq p}={\rm diag}\{(\hat{\bm{\Sigma}}(t_i)+\rho\bm{I})_{(-u,-u)}\}_{1\leq u\leq p},
	$$
	where $\hat{\bm{\Sigma}}(t_i)$ is the kernel estimate of the covariance matrix as in Section \ref{subsec:loggle}. That is, $\mathbf{X}(t_i)^T\mathbf{X}(t_i)+\rho\bm{I}$ is a block diagonal matrix with $p$ blocks, where the $u$th block, the $(p-1)\times(p-1)$ matrix $(\hat{\bm{\Sigma}}(t_i)+\rho\bm{I})_{(-u,-u)}$, is the matrix $\hat{\bm{\Sigma}}(t_i)+\rho\bm{I}$ with the $u$th row and the $u$th column deleted.
	
	Moreover, 
	\begin{align*}
	\mathbf{X}(t_i)^T\mathbf{Y}(t_i)=&((\tilde{\mathbf{X}}_{(-1)}(t_i)^T\tilde{\bm{X}}_1(t_i))^T,\ldots, (\tilde{\mathbf{X}}_{(-p)}(t_i)^T\tilde{\bm{X}}_p(t_i))^T)^T\\
	=&((\hat{\bm{\Sigma}}(t_i)_{(-1,1)})^T,\ldots,(\hat{\bm{\Sigma}}(t_i)_{(-p,p)})^T)^T.
	\end{align*}
	That is, $\mathbf{X}(t_i)^T\mathbf{Y}(t_i)$ is a $p(p-1)\times1$ column vector consisting of $p$ sub-vectors, where the $u$th sub-vector is the $u$th column of $\hat{\bm{\Sigma}}(t_i)$ with the $u$th element (i.e., the diagonal element) deleted.
	
	Since $\mathbf{X}(t_i)^T\mathbf{X}(t_i)+\rho\bm{I}$ and $\mathbf{X}(t_i)^T\mathbf{Y}(t_i)$ can be decomposed into blocks, $\bm{\beta}^s(t_i)$ can be solved block-wisely: 
	$$
	\bm{\beta}_u^s(t_i)=((\hat{\bm{\Sigma}}(t_i)+\rho\bm{I})_{(-u,-u)})^{-1}(\hat{\bm{\Sigma}}(t_i)_{(-u,u)}+\rho(\bm{Z}_u^{s-1}(t_i)-\bm{U}_u^{s-1}(t_i))),\ u=1,\ldots,p,
	$$
	where $\bm{\beta}_u^s(t_i)=(\beta_{u1}^s(t_i),\ldots,\beta_{u,u-1}^s(t_i),\ldots,\beta_{u,u+1}^s(t_i),\ldots,\beta_{up}^s(t_i))$ is a $(p-1)\times1$ column vector, $\bm{\beta}^{s}(t_i)=(\bm{\beta}_1^s(t_i)^T,\ldots,\bm{\beta}_p^s(t_i)^T)^T$, and $\bm{Z}_u^{s-1}(t_i)=(Z_{u1}^{s-1}(t_i),\ldots,Z_{up}^{s-1}(t_i))$ and $\bm{U}_u^{s-1}(t_i)=(U_{u1}^{s-1}(t_i),\ldots,U_{up}^{s-1}(t_i))$ contain the corresponding elements in $\bm{Z}^{s-1}(t_i)$ and $\bm{U}^{s-1}(t_i)$, respectively.
	
	Here, we need to solve $p$ linear systems, each with $p$ equations. One way is to conduct Cholesky decompositions of  the matrices $(\hat{\bm{\Sigma}}(t_i)+\rho\bm{I})_{(-u,-u)}$, $u=1,\ldots,p$ in advance and use Gaussian elimination to solve the corresponding triangular linear systems. To do this, we apply Cholesky decomposition to $\hat{\bm{\Sigma}}(t_i)+\rho\bm{I}$ followed by $p$ Givens rotations. This has overall time complexity $O(p^3)$, the same as the time complexity of the subsequent $p$ applications of Gaussian eliminations. Note that, if we had performed Cholesky decomposition on each of the  $(p-1)\times (p-1)$ matrix directly, the total time complexity would have been $O(p^4)$. 
		The details of conducting Cholesky decompositions of  the matrices $(\hat{\bm{\Sigma}}(t_i)+\rho\bm{I})_{(-u,-u)} (u=1,\ldots,p)$ through Givens rotations are given in \ref{sec:givens_rotation} of the Supplementary Material.

	\item
	$$\mathbb{Z}_k^{s}=\arg\min_{\mathbb{Z}_k}\left[\frac{\rho}{2}\sum_{i\in \mathcal{N}_{k,d}}||\bm{Z}(t_i)-\bm{\beta}^{s}(t_i)-\bm{U}^{s-1}(t_i)||_2^2+\lambda\sum_{u<v}\sqrt{\sum_{i\in \mathcal{N}_{k,d}}\left[Z_{uv}(t_i)^2+Z_{vu}(t_i)^2\right]}\right].$$
	For $i\in \mathcal{N}_{k,d}$, $1\leq u\neq v\leq p$, it is easy to see that
	\begin{align*}
	Z_{uv}^{s}(t_i)=&\left(1-\frac{\lambda}{\rho\sqrt{\sum_{j\in \mathcal{N}_{k,d}}\left[(\beta_{uv}^{s}(t_j)+U_{uv}^{s-1}(t_j))^2+(\beta_{vu}^{s}(t_j)+U_{vu}^{s-1}(t_j))^2\right]}}\right)_+\\
	&\cdot(\beta_{uv}^{s}(t_i)+U_{uv}^{s-1}(t_i)).
	\end{align*}
	\item
	For $i\in \mathcal{N}_{k,d}$,
	$$\bm{U}^{s}(t_i)=\bm{U}^{s-1}(t_i)+\bm{\beta}^{s}(t_i)-\bm{Z}^{s}(t_i).$$
\end{enumerate}

\subsubsection*{Over-relaxation}
In steps (ii) and (iii), we  replace $\bm{\beta}^s(t_i)$ by $\alpha\bm{\beta}^s(t_i)+(1-\alpha)\bm{Z}^{s-1}(t_i)$, where the relaxation parameter $\alpha$ is set to be 1.5. It is suggested in \cite{boyd2011distributed} that over-relaxation with $\alpha\in[1.5, 1.8]$ can improve convergence.

\subsubsection*{Stopping criterion}

 The norm of the primal residual at step $s$ is 
$||\bm{r}^s||_2=\sqrt{\sum_{i\in \mathcal{N}_{k,d}}||\bm{\beta}^s(t_i)-\bm{Z}^s(t_i)||_2^2}$,  
and the norm of the dual residual at step $s$ is
$||\bm{d}^s||_2=\sqrt{\sum_{i\in \mathcal{N}_{k,d}}||\bm{Z}^s(t_i)-\bm{Z}^{s-1}(t_i)||_2^2}$. Define the feasibility tolerance for the primal as  $\epsilon^{pri}=\epsilon^{abs}\sqrt{p|\mathcal{N}_{k,d}|}+\epsilon^{rel}\max\{\sqrt{\sum_{i\in \mathcal{N}_{k,d}}||\bm{\beta}^s(t_i)||_2^2},$ $\sqrt{\sum_{i\in \mathcal{N}_{k,d}}||\bm{Z}^s(t_i)||_2^2}\}$, 
and the feasibility tolerance for the dual as  $\epsilon^{dual}=\epsilon^{abs}\sqrt{p|\mathcal{N}_{k,d}|}+\epsilon^{rel}\sqrt{\sum_{i\in \mathcal{N}_{k,d}}||\bm{U}^s(t_i)||_2^2}$. 
Here $\epsilon^{abs}$ is the absolute tolerance and in practice is often set as $10^{-5}$ or $10^{-4}$, and $\epsilon^{rel}$ is the relative tolerance and in practice is often set as $10^{-3}$ or $10^{-2}$.
The stopping criterion is that the algorithm stops if and only if $||\bm{r}^s||_2\leq\epsilon^{pri}$ and $||\bm{d}^s||_2\leq\epsilon^{dual}$.

\bibliographystyle{agsm}

\bibliography{loggle_reference}

\clearpage
\newpage
\setcounter{page}{1}
\setcounter{section}{0}
\renewcommand{\thesection}{S.\arabic{section}}
\setcounter{subsection}{0}
\renewcommand{\thesubsection}{S.\arabic{section}.\arabic{subsection}}
\setcounter{equation}{0}
\renewcommand{\theequation}{S.\arabic{equation}}
\setcounter{figure}{0}
\renewcommand{\thefigure}{S.\arabic{figure}}
\setcounter{table}{0}
\renewcommand{\thetable}{S.\arabic{table}}
\setcounter{proposition}{0}
\renewcommand{\theproposition}{S.\arabic{proposition}}
\setcounter{lemma}{0}
\renewcommand{\thelemma}{S.\arabic{lemma}}
\setcounter{corollary}{0}
\renewcommand{\thecorollary}{S.\arabic{corollary}}

\begin{center}
	{\large\bf Estimating Time-Varying Graphical Models\\
	SUPPLEMENTARY MATERIAL}
\end{center}

\section{Algorithm details} 

\subsection{ADMM algorithm in likelihood-based \texttt{loggle}}\label{sec:ADMM}

To solve the optimization problem in (\ref{eq:loggle}) using ADMM algorithm, we notice that the problem can be written as
\begin{align*}
&\textrm{minimize }_{\mathbb{\Omega}_k,\mathbb{Z}_k}\sum_{i\in \mathcal{N}_{k,d}}\left[\text{tr}\left(\bm{\Omega}(t_i)\hat{\bm{\Sigma}}(t_i)\right)-\log|\bm{\Omega}(t_i)|\right]
+\lambda\sum_{u\neq v}\sqrt{\sum_{i\in \mathcal{N}_{k,d}}Z_{uv}(t_i)^2},\\
&\textrm{subject to }\bm{\Omega}(t_i)-\bm{Z}(t_i)=\bm{0},\;\bm{\Omega}(t_i)\succ\bm{0},\;i\in \mathcal{N}_{k,d},
\end{align*}
where $\mathcal{N}_{k,d}=\{i:|t_i-t_k|\leq d\}$, $\mathbb{\Omega}_k=\{\bm{\Omega}(t_i)\}_{i\in \mathcal{N}_{k,d}}$ and $\mathbb{Z}_k=\{\bm{Z}(t_i)\}_{i\in \mathcal{N}_{k,d}}$. Note $\bm{\Omega}(t_i)=(\Omega_{uv}(t_i))$ and  $\bm{Z}(t_i)=(Z_{uv}(t_i))$  are $\mathbb{R}^{p\times p}$ matrices.

The scaled augmented Lagrangian is
\begin{align*}
L_\rho(\mathbb{\Omega}_k,\mathbb{Z}_k,\mathbb{U}_k)=&\sum_{i\in \mathcal{N}_{k,d}}\left[\text{tr}\left(\bm{\Omega}(t_i)\hat{\bm{\Sigma}}(t_i)\right)-\log|\bm{\Omega}(t_i)|\right]
+\lambda\sum_{u\neq v}\sqrt{\sum_{i\in \mathcal{N}_{k,d}}Z_{uv}(t_i)^2}\\
&+\frac{\rho}{2}\sum_{i\in \mathcal{N}_{k,d}}||\bm{\Omega}(t_i)-\bm{Z}(t_i)+\bm{U}(t_i)||_F^2,
\end{align*}
where $\mathbb{U}_k=\{\bm{U}(t_i)\}_{i\in \mathcal{N}_{k,d}}$ are dual variables ($\bm{U}(t_i)=(U_{uv}(t_i))\in\mathbb{R}^{p\times p}$).

The ADMM algorithm is as follows. We first initialize $\bm{Z}^{(0)}(t_i)=\bm{0}$, $\bm{U}^{(0)}(t_i)=\bm{0}$, $i\in \mathcal{N}_{k,d}$. We also need to specify $\rho (>0)$, which in practice is recommended to be  $\approx\lambda$ \citep{wahlberg2012admm}. For step $s=1,2,\ldots$ until convergence:
\begin{enumerate}[(i)]
	\item
	For $i\in \mathcal{N}_{k,d}$,
	\begin{align*}
	\bm{\Omega}^{s}(t_i)=\arg\min_{\bm{\Omega}(t_i)\succ\bm{0}}\left[\text{tr}\left(\bm{\Omega}(t_i)\hat{\bm{\Sigma}}(t_i)\right)-\log|\bm{\Omega}(t_i)|+\frac{\rho}{2}||\bm{\Omega}(t_i)-\bm{Z}^{s-1}(t_i)+\bm{U}^{s-1}(t_i)||_F^2\right].
	\end{align*}
	Set the derivative to be 0, we have
	$$\hat{\bm{\Sigma}}(t_i)-\rho(\bm{Z}^{s-1}(t_i)-\bm{U}^{s-1}(t_i))=\bm{\Omega}^{-1}(t_i)-\rho\bm{\Omega}(t_i).$$
	Let $\bm{Q\Lambda Q}^T$ denote the eigen-decomposition of $\hat{\bm{\Sigma}}(t_i)-\rho(\bm{Z}^{s-1}(t_i)-\bm{U}^{s-1}(t_i))$, where $\bm{\Lambda}={\rm diag}\{\lambda_1,\ldots,\lambda_p\}$, then
	$$\bm{\Omega}^{s}(t_i)=\bm{Q}\tilde{\bm{\Lambda}}\bm{Q}^T,$$ where $\tilde{\bm{\Lambda}}$ is the diagonal matrix with $j$th diagonal element $\frac{-\lambda_j+\sqrt{\lambda_j^2+4\rho}}{2\rho}$.
	\item
	$$\mathbb{Z}_k^{s}=\arg\min_{\mathbb{Z}_k}\left[\frac{\rho}{2}\sum_{i\in \mathcal{N}_{k,d}}||\bm{Z}(t_i)-\bm{\Omega}^{s}(t_i)-\bm{U}^{s-1}(t_i)||_F^2+\lambda\sum_{u\neq v}\sqrt{\sum_{i\in \mathcal{N}_{k,d}}Z_{uv}(t_i)^2}\right].$$
	For $i\in \mathcal{N}_{k,d}$, it is easy to see that the diagonal elements
	$$Z_{uu}^{s}(t_i)=\Omega_{uu}^{s}(t_i)+U_{uu}^{s-1}(t_i),\ u=1,\ldots,p.$$
	For the off-diagonal elements, one can show that they should take the form
	$$Z_{uv}^{s}(t_i)=\left(1-\frac{\lambda}{\rho\sqrt{\sum_{j\in \mathcal{N}_{k,d}}(\Omega_{uv}^{s}(t_j)+U_{uv}^{s-1}(t_j))^2}}\right)_+(\Omega_{uv}^{s}(t_i)+U_{uv}^{s-1}(t_i)).$$
	\item
	For $i\in \mathcal{N}_{k,d}$,
	$$\bm{U}^{s}(t_i)=\bm{U}^{s-1}(t_i)+\bm{\Omega}^{s}(t_i)-\bm{Z}^{s}(t_i).$$
\end{enumerate}
Note that in Step (i), the positive-definiteness constraint on $\{\bm{\Omega}(t_i)\}_{i\in \mathcal{N}_{k,d}}$ is automatically satisfied by implementing eigen-decomposition.

\subsubsection*{Over-relaxation}
In step (ii) and (iii), we replace $\bm{\Omega}^s(t_i)$ by $\alpha\bm{\Omega}^s(t_i)+(1-\alpha)\bm{Z}^{s-1}(t_i)$, where the relaxation parameter $\alpha$ is set to be 1.5. 

\subsubsection*{Stopping criterion}
The norm of the primal residual at step $s$ is $||\bm{r}^s||_F=\sqrt{\sum_{i\in \mathcal{N}_{k,d}}||\bm{\Omega}^s(t_i)-\bm{Z}^s(t_i)||_F^2}$, and the norm of the dual residual at step $s$ is $||\bm{d}^s||_F=\sqrt{\sum_{i\in \mathcal{N}_{k,d}}||\bm{Z}^s(t_i)-\bm{Z}^{s-1}(t_i)||_F^2}$. Define the feasibility tolerance for the primal as $\epsilon^{pri}=\epsilon^{abs}\sqrt{p|\mathcal{N}_{k,d}|}+\epsilon^{rel}\max\{\sqrt{\sum_{i\in \mathcal{N}_{k,d}}||\bm{\Omega}^s(t_i)||_F^2},$ $\sqrt{\sum_{i\in \mathcal{N}_{k,d}}||\bm{Z}^s(t_i)||_F^2}\}$, and the feasibility tolerance for the dual as $\epsilon^{dual}=\epsilon^{abs}\sqrt{p|\mathcal{N}_{k,d}|}+\epsilon^{rel}\sqrt{\sum_{i\in \mathcal{N}_{k,d}}||\bm{U}^s(t_i)||_F^2}$.
Here $\epsilon^{abs}$ is the absolute tolerance and in practice is often set as $10^{-5}$ or $10^{-4}$, and $\epsilon^{rel}$ is the relative tolerance and in practice is often set as $10^{-3}$ or $10^{-2}$. The stopping criterion is that the algorithm stops if and only if $||\bm{r}^s||_F\leq\epsilon^{pri}$ and $||\bm{d}^s||_F\leq\epsilon^{dual}$.

\subsection{ADMM algorithm in pseudo-likelihood-based \texttt{loggle}: additional details}
\subsubsection*{Givens rotation}\label{sec:givens_rotation}
Let $\bm{A}=\hat{\bm{\Sigma}}(t_i)+\rho\bm{I}$, a $p\times p$ positive definite matrix. We aim to efficiently obtain the Cholesky decomposition of $\bm{A}_j$: $\bm{A}_j=\bm{U}_j^T\bm{U}_j$, where $\bm{A}_j$ is a $(p-1)\times(p-1)$ matrix obtained by deleting the $j$th row and the $j$th column of $\bm{A}$, and $\bm{U}_j$ is an upper triangular matrix.

We first apply the Cholesky decomposition to $\bm{A}$: $\bm{A}=\bm{U}^T\bm{U}$, where $\bm{U}$ is an upper triangular matrix. It is easy to see that $\bm{A}_j=\tilde{\bm{U}}_j^T\tilde{\bm{U}}_j$, where $\tilde{\bm{U}}_j$ is a $p\times(p-1)$ matrix from deleting the $j$th column of $\bm{U}$. We then apply the Givens rotation to $\tilde{\bm{U}}_j$ to get its QR decomposition: $\tilde{\bm{U}}_j=\bm{Q}_j\bm{R}_j$, where $\bm{Q}_j$ is a $p\times p$ orthogonal matrix and $\bm{R}_j^T=[\bm{U}_j^T, \bm{0}]$ where $\bm{U}_j$ is a $(p-1)\times(p-1)$ upper triangular matrix. Thus $\bm{A}_j=(\bm{Q}_j\bm{R}_j)^T\bm{Q}_j\bm{R}_j=\bm{R}_j^T\bm{R}_j=\bm{U}_j^T\bm{U}_j$ is exactly the Cholesky decomposition of $\bm{A}_j$.

Note that the cost of Cholesky decomposition is $O(p^3)$ and the cost of QR decomposition by Givens rotation is $O(p^2)$, hence by using the Givens rotation, the total computational complexity for implementing $p$ Cholesky decompositions is $O(p^3)$.

\subsection{De-trend}
\label{sec:detrend}
Our observations $\{\bm{x}_k\}_{k=1,\ldots,N}$ are drawn from $\mathcal{N}_p(\bm{\mu}(t_k),\bm{\Sigma}(t_k))$ ($k=1,\ldots,N$) independently. For simplicity, we can assume that these observations are centered so that each $\bm{x}_k$ is drawn independently from $\mathcal{N}_p(\bm{0},\bm{\Sigma}(t_k))$ ($k=1,\ldots,N$). To achieve this, we first obtain a kernel estimate of the mean function $\hat{\bm{\mu}}(t)=\sum_{j=1}^N\omega_h^{t_j}(t)\bm{x}_j$ by using R package \texttt{sm}, where $\omega_h^{t_j}(t)=\frac{K_h(t_j-t)}{\sum_{j=1}^NK_h(t_j-t)}$ and $K_h(\cdot)$ is a normal kernel function with $h$ being its standard deviation. We then subtract the kernel estimated mean $\hat{\bm{\mu}}(t_k)$ from $\bm{x}_k$ for each $k=1,\ldots,N$.

\section{Model tuning: additional details}
\label{sec:supp_tuning}
\subsection{Early stopping in grid search}\label{sec:thresh_grid}
When the number of nodes  $p$ is large, it is often very time consuming to estimate a dense graph. Moreover, when sample size is relatively small, dense models  seldom lead to good cross-validation scores (so they will not be selected anyway). Thus,  we will fit models for a decreasing sequence of 
$\lambda$ and when the edge number of the fitted graph at a $\lambda$ exceeds a prespecified threshold (e.g, 5 times of $p$), we will stop the grid search as  smaller values of $\lambda$ usually lead to even denser fitted graphs.

\subsection{Coarse grid search}\label{sec:coarse_grid}
To further reduce the computational cost, we implement a coarse grid search followed by a fine grid search for model tuning. Specifically, we first apply cross-validation on coarse grids to search for appropriate values of $h$, and then we  use finer grids to search for good values of $d$ and $\lambda$.


\section{Simulation: additional details}\label{sec:simu_add}
\begin{figure}[H]
	\begin{center}
		\begin{tabular}{cc}
			
			\includegraphics[width = 2.8in]{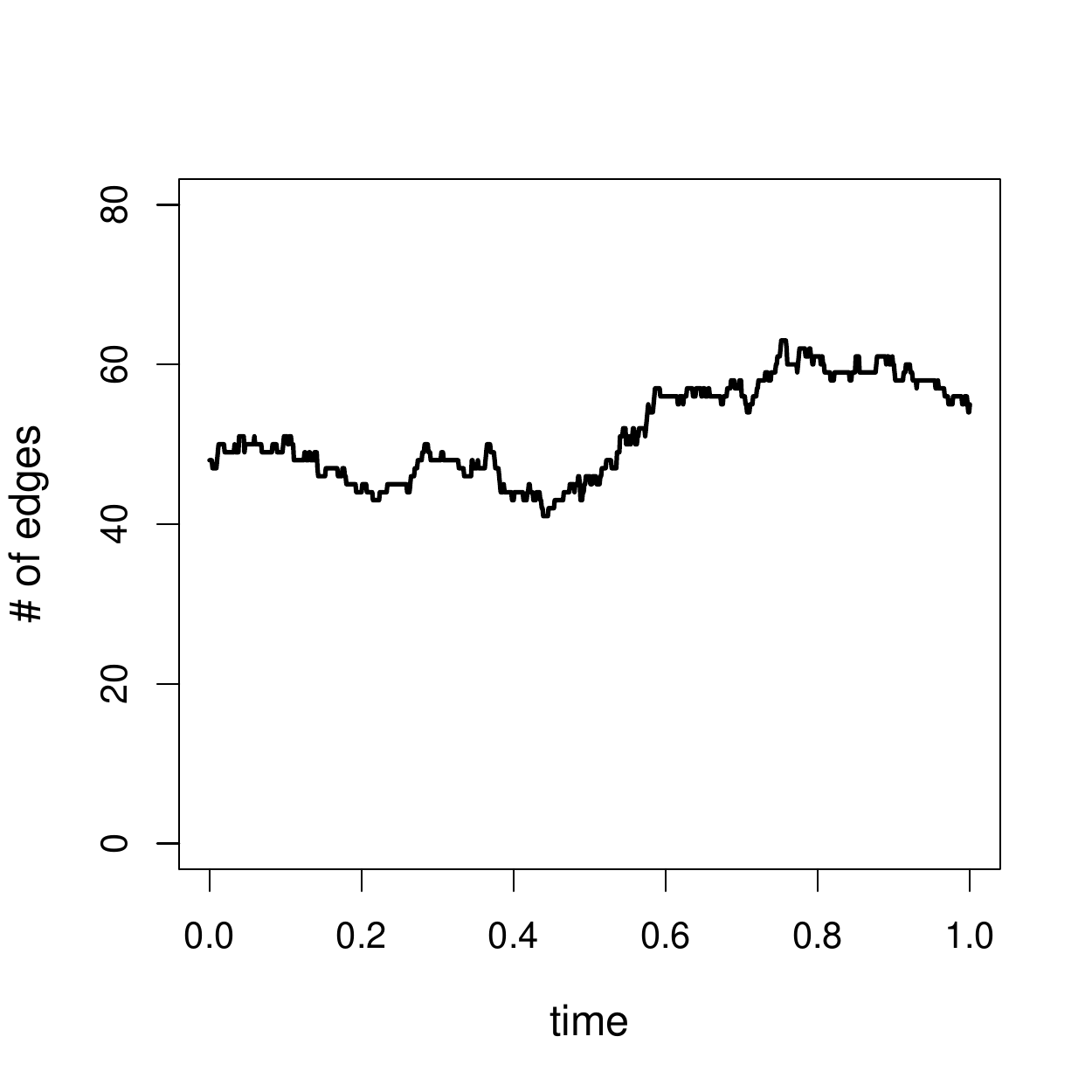} & \includegraphics[width = 2.8in]{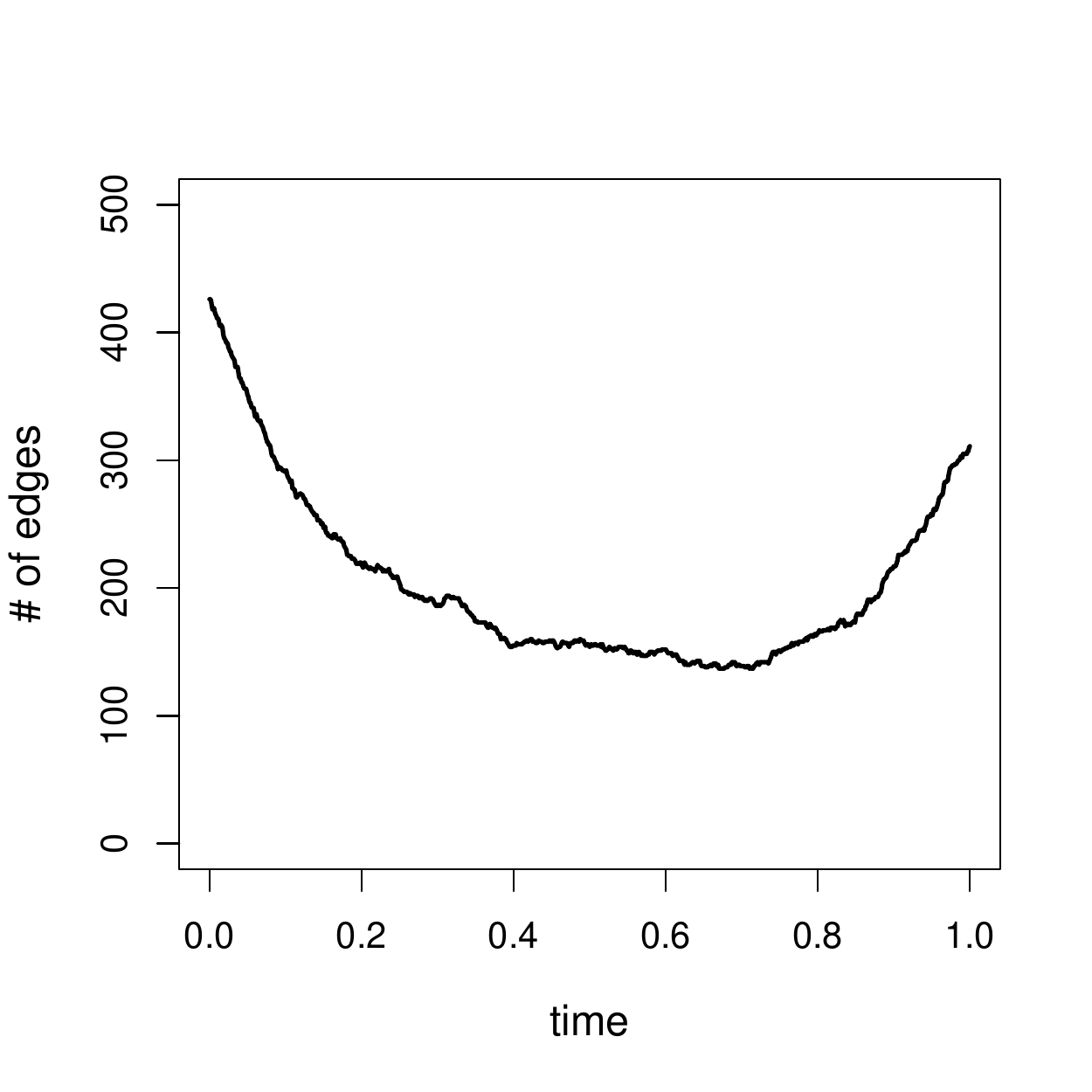}\\
			
		\end{tabular}
	\end{center}
	\caption{\textbf{Simulation.} Number of edges versus time. Left: $p=100$ time-varying graphs model; Right: $p=500$ time-varying graphs model.}\label{fig:ed_num}
\end{figure}

\section{S\&P 500 Stock Price: additional details}\label{sec:real_data_add}

\begin{figure}[H]
	\begin{center}
		\includegraphics[width = 3in]{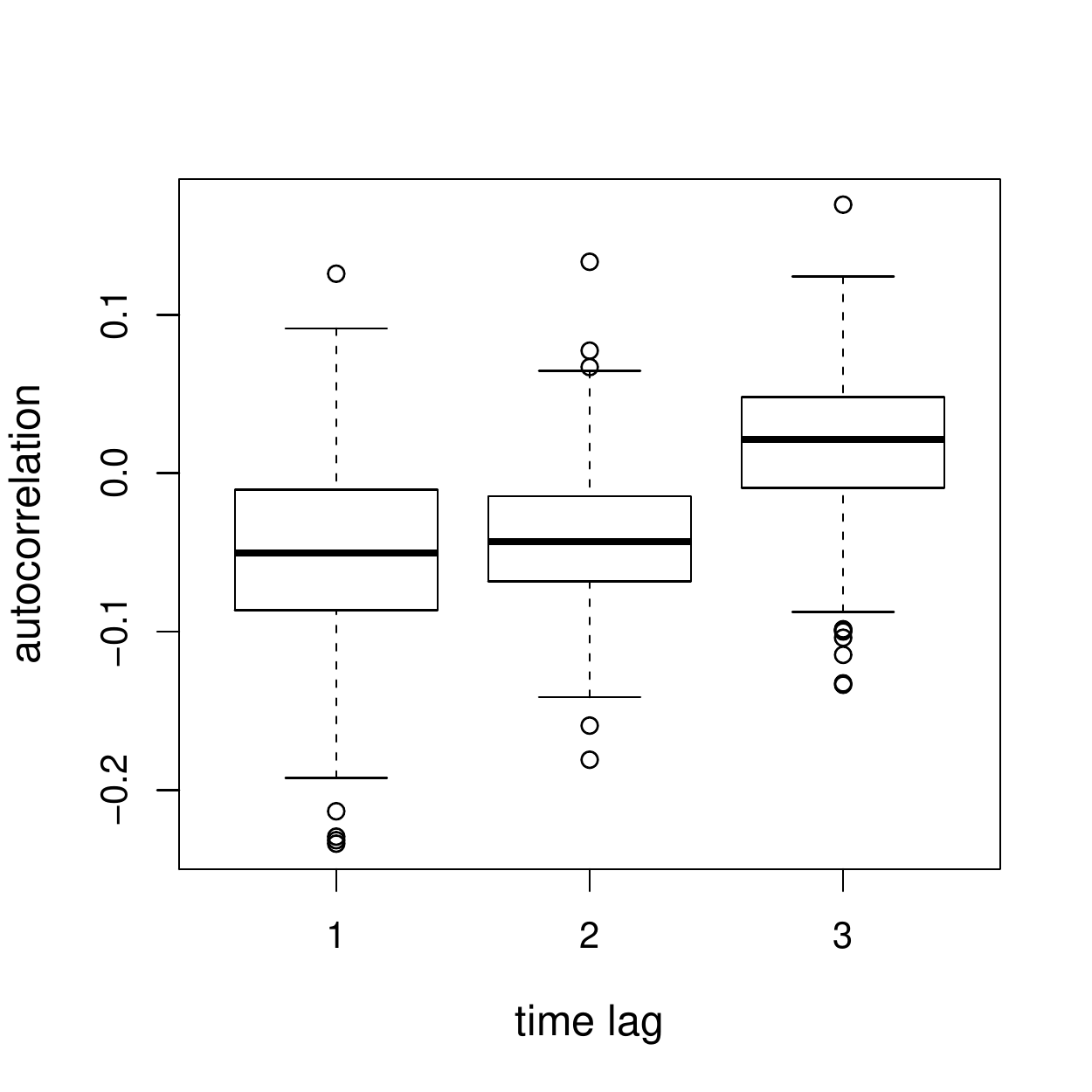}\\
	\end{center}
	\caption{Boxplots of autocorrelations at time lag 1, 2 and 3.}\label{fig:acf}
\end{figure}

\end{document}